\documentclass[10pt,twocolumn,letterpaper]{article}

\usepackage[accsupp]{axessibility}

\usepackage[pagenumbers]{cvpr} 

\usepackage{graphicx}
\usepackage{amsmath}
\usepackage{amssymb}
\usepackage{booktabs}
\usepackage{float}
\usepackage{multirow}
\usepackage{graphicx}
\usepackage{stmaryrd}
\usepackage{algorithm}
\usepackage{algpseudocode}
\usepackage{xspace}
\usepackage[dvipsnames]{xcolor}
\usepackage{annotate-equations}
\usepackage{duckuments}
\usepackage{tikz}
\usepackage{mathtools}

\usepackage[pagebackref,breaklinks,colorlinks]{hyperref}

\usepackage[capitalize]{cleveref}
\crefname{section}{Sec.}{Secs.}
\Crefname{section}{Section}{Sections}
\Crefname{table}{Table}{Tables}
\crefname{table}{Tab.}{Tabs.}

\def\cvprPaperID{24} 
\def\confName{CVPR}
\def\confYear{2023}

\definecolor{applegreen}{rgb}{0.55, 0.71, 0.0}

\newcommand{\task}{DASS\xspace}

\newcommand{\method}{CONFETI\xspace}

\definecolor{protoCol}{rgb}{0.99, 0.75, 0.75}
\definecolor{darkgreen}{RGB}{119,185,0}

\begin{document}

\title{Contrast, Stylize and Adapt: Unsupervised Contrastive Learning \\Framework for Domain Adaptive Semantic Segmentation}

 \author{Tianyu Li$^{1,2}$, Subhankar Roy$^{2}$, Huayi Zhou$^{1}$, Hongtao Lu$^{1}$,  Stéphane Lathuilière$^{2}$   \\
  $^{1}$ Shanghai Jiao Tong University, Shanghai \hspace{0.5cm} $^{2}$ LTCI, Télécom-Paris, Institut Polytechnique de Paris \\
 {\tt\small hugo\_li@sjtu.edu.cn}
 }

\maketitle

\begin{abstract}
   To overcome the domain gap between synthetic and real-world datasets, unsupervised domain adaptation methods have been proposed for semantic segmentation. Majority of the previous approaches have attempted to reduce the gap either at the pixel or feature level, disregarding the fact that the two components interact positively. To address this, we present \textbf{CON}trastive \textbf{FE}a\textbf{T}ure and p\textbf{I}xel alignment (CONFETI) for bridging the domain gap at both the pixel and feature levels using a unique contrastive formulation. We introduce well-estimated prototypes by including category-wise cross-domain information to link the two alignments: the pixel-level alignment is achieved using the jointly trained style transfer module with the \textbf{prototypical semantic consistency}, while the feature-level alignment is enforced to cross-domain features with the \textbf{pixel-to-prototype contrast}. Our extensive experiments demonstrate that our method outperforms existing state-of-the-art methods using DeepLabV2. Our code\footnote{Code: \url{https://github.com/cxa9264/CONFETI}} has been made publicly available.
   
\end{abstract}

\section{Introduction}
\label{sec:intro}
Semantic segmentation is a fundamental task in computer vision that consists in predicting the class label of each pixel in an image~\cite{csurka2022semantic}. Segmentation has been the focus of extensive research in the supervised regime, leading to considerable progress in recent years~\cite{badrinarayanan2017segnet, deeplab, chen2017rethinking,seformer}. Much of this progress can be attributed to the availability of large-scale annotated datasets, such as Cityscapes~\cite{cs} and ADE20K~\cite{zhou2017scene}. However, the cost of manual annotation often compels the practitioners to rely on pre-trained models in test environments, without fine-tuning. Unfortunately, these pre-trained models generally perform poorly on test samples that differ from the training data, due to the so-called \textit{domain shift} problem~\cite{torralba2011unbiased}. To address this problem, Domain Adaptive Semantic Segmentation (\task) methods~\cite{csurka2021unsupervised} have been proposed that enable learning on the domain of interest, without needing annotations.

\begin{figure}
    \centering
    \includegraphics[width=\columnwidth]{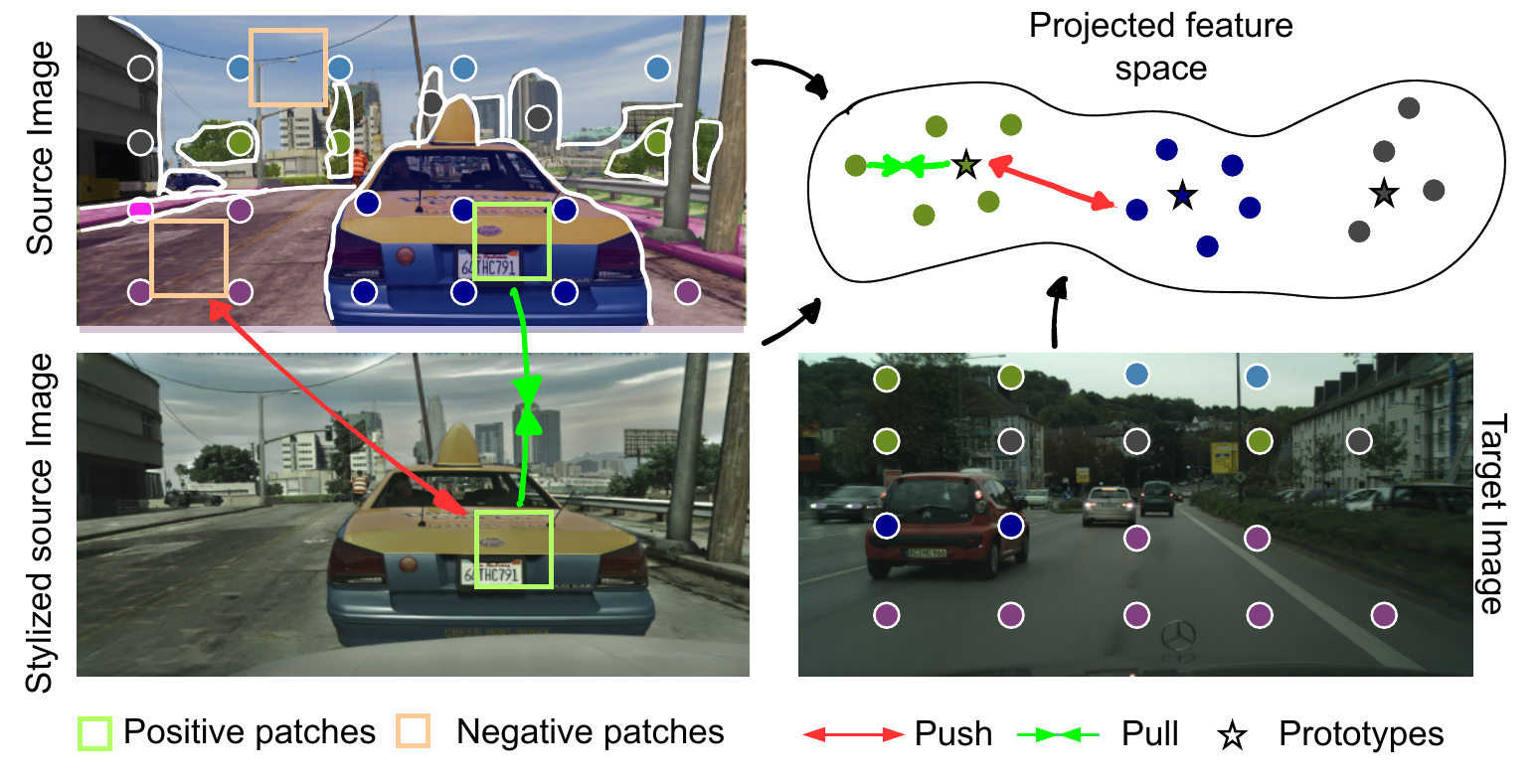}
    \vspace{-5mm}
    \caption{They key idea of our proposed \textbf{\method} is to use contrastive learning to unify \textit{feature-level alignment} with \textit{pixel-level alignment}. Features of the pixels from the same class, but across domains, are pulled towards it's corresponding prototype and pushed apart from dissimilar ones. For improved style transfer it enforces that the positive patches in source and stylized images are closer than the negative patches in the projected feature space.}
    \label{fig:teaser}
    \vspace{-5mm}
\end{figure}

Traditionally, the \task methods have been designed to address the problem primarily from one of the two fronts: feature-level alignment~\cite{huang2018domain,hong2018conditional} or pixel-level alignment~\cite{chen2019crdoco,murez2018image}, both aiming to align the labelled source and unlabelled target domain. Very recently, self-training~\cite{araslanov2021self,daformer} with student-teacher framework~\cite{mt} has emerged as an effective technique to iteratively fine-tune on the target domain by using the most confident pseudo-labels. With so many genres of existing methods for DASS, it begs the question: \textcolor{black}{\textit{\textbf{How to combine the best of the worlds in \task?}}}

To find answer to this question we turn our attention to the contrastive formulation, InfoNCE~\cite{gutmann2010noise}, that has been found to be effective for a myriad of tasks such as supervised segmentation~\cite{wang2021exploring}, weakly supervised segmentation~\cite{wsss} and unpaired image translation~\cite{cut}, among others. Given the versatility of the contrastive loss in addressing representation learning and unpaired image translation, both of which have proven to be useful for \task~\cite{cycada}, in this work we propose an unsupervised contrastive learning framework for \task. We leverage contrastive learning to conduct
both \textit{feature-level} and \textit{pixel-level} alignment, while synergistically using the mean-teacher framework.

From the perspective of feature-level alignment, the contrastive loss ensures that the representation of pixels belonging to the same class, but \textit{across} domains, are closer to each other in an embedding space (\ie, intra-class compactness) while being discriminative to other unrelated classes (\ie, inter-class dispersion). Such a formulation comes with two key advantages: (\textbf{i}) it enables us to contrast with pixel locations not only from the same image but from other images (both source and target domain); and (\textbf{ii}) it allows to consider the global structure present in a scene, which is in sharp contrast to methods (for \eg, self-training) that treat each pixel individually. To reduce computation, we maintain classwise \textit{prototypes} computed from Class Activation Maps~\cite{zhou2016learning} (see Sec.~\ref{sec:feat-align} for details), and enforce \textit{pixel-to-prototype} contrast where the pixel embeddings are contrasted with the prototypes instead of pixels.

On the other front of pixel-level alignment, which essentially consists in generating target-\textit{like} source images, the contrastive learning helps in the style transfer by making unpaired image translation one-sided~\cite{cut}, instead of the classical bi-directional cycle-consistent translation~\cite{zhu2017unpaired}. Concretely, we adopt CUT~\cite{cut} that uses a patchwise contrastive loss to ensure that the feature representation of corresponding patches in the source and target-\textit{like} (or stylized) source image are closer in the embedding space than other random patches. To further improve the stylization, we propose to use a semantic consistency loss that makes sure that the semantic content is not altered during the stylization process (see Sec.~\ref{sec:pixel-align} for details).

We call our framework \textbf{CON}trastive \textbf{FE}a\textbf{T}ure and p\textbf{I}xel alignment (\method) as it allows to amalgamate both feature-level and pixel-level alignment using the unique formulation of contrastive loss (see Fig.~\ref{fig:teaser}). We also show that \method can be seamlessly integrated with the mean-teacher framework, where the prototypes are computed using the teacher network, and the student network learns to match the representation of the corresponding prototype.

In summary, our \textbf{contributions} are three-fold: (\textbf{i}) We propose an unsupervised contrastive learning framework called \method that enables both feature-level and pixel-level alignment for addressing \task; (\textbf{ii}) We show that \method can easily be integrated with the very effective self-training strategy; and (\textbf{iii}) We extensively evaluate our method on standard \task benchmarks and set new state-of-the-art results when compared with methods that use the common DeepLab~\cite{deeplab} segmentation network.

\section{Related Works}
\label{sec:related}
\noindent\textbf{Domain Alignment in \task.} Following the success of domain \textit{alignment} in image classification, the semantic segmentation methods have adopted various alignment techniques, which ensure that the source and target distributions are aligned at different levels of the pipeline under some metric. Particular to \task, the three levels are namely latent feature space, input (or pixel) space and output space. First, the feature-level alignment \task methods seek to minimize the distance between the marginal feature distributions of the source and the target, by either minimizing Maximum Mean Discrepancy along with aligning the correlation matrices~\cite{bermudez2018domain}, or by using a domain discriminator to increase \textit{domain confusion} in the learned features~\cite{huang2018domain,hoffman2016fcns,shen2019regularizing,vu2019advent,luo2019taking}. Second, the pixel-level alignment consists in bridging the domain gap via style transfer~\cite{adain,zhu2017unpaired}, which involves transferring the `appearance' of the target domain onto the source images. The \task methods that incorporate pixel-level alignment~\cite{pizzati2020domain,choi2019self,chen2019crdoco,murez2018image,yang2020phase,fda} have proven to be very effective since the content do not change drastically in the \task benchmarks. Third, the output-level alignment methods circumvent the high dimensionality of the latent feature space and instead perform adversarial adaptation in the output space of the network~\cite{tsai2019domain,pan2020unsupervised,vu2019advent}. The complementary nature of the alignment techniques has led to the development of \task methods~\cite{bidir,cycada,toldo2020unsupervised} that combine different domain alignments, to better mitigate the domain shift. Our proposed \method also harmoniously combines feature-level alignment with pixel-level alignment, but via contrastive learning~\cite{gutmann2010noise}.

\vspace{1mm}

\noindent\textbf{Self-training in \task.} Drawing inspirations from the semi-supervised learning, the idea of using pseudo-labels generated for unlabelled target data, and using them to iteratively fine-tune (or self-train) the target model is also prevalent in \task~\cite{zou2018unsupervised,zheng2021rectifying,zhang2023cooperative}. Several of the very recent \task methods using the self-training strategy have adopted the popular idea of model \textit{ensembling} for obtaining pseudo-labels. In particular, these methods~\cite{araslanov2021self,daformer,choi2019self} use the mean-teacher~\cite{mt} (or student teacher) framework where the teacher network, which is an exponential moving average of the student network weights, provides pseudo-labels to train on the target data. In our work we also utilize the teacher network to obtain pseudo-labels, which are then used by the student network for the feature-level alignment.

\vspace{1mm}

\noindent\textbf{Contrastive learning in \task.} Learning discriminative visual features in a self-supervised manner, where given an anchor data point, the network must distinguish a similar sample from other dissimilar samples, forms the core idea of contrastive learning~\cite{gutmann2010noise,chen2020simple,he2020momentum}. Due to its effectiveness, \task methods~\cite{melas2021pixmatch, marsden2022contrastive, proca, sepico} have adopted it for learning compact latent embedding space. For instance, CLST~\cite{marsden2022contrastive} leverages self-training to obtain pseudo-labels for computing class-specific prototypes (or centroids), which are then used in a contrastive manner to learn more compact features. Similarly, ProCA~\cite{proca} contrasts the pixel representation with the prototypes, except the source prototypes are updated with the target in a moving average fashion. SePiCo~\cite{sepico} goes a step further and estimates the distribution
of each prototypes, rather than point estimates. Different from the previous works that exploit contrastive learning, our \method computes the prototypes from mixed images, obtained with ClassMix~\cite{dacs}, and the class activation maps~\cite{zhou2016learning}. Besides feature-level alignment, we also employ contrastive learning for the pixel-level alignment.

\section{Methods}
\label{sec:methods}
In this work we propose \textbf{CON}trastive \textbf{FE}a\textbf{T}ure and p\textbf{I}xel alignment (\method), a domain adaptive semantic segmentation (DASS) method, that leverages the contrastive formulation to (\textbf{i}) learn a well structured pixel embedding space for feature-level alignment; and (\textbf{ii}) foster more accurate style transfer between the source and the target for pixel-level alignment. Before we introduce our method, we formalize the problem and discuss the preliminaries.

\subsection{Preliminaries}
\label{sec:prel}
\noindent\textbf{Problem Definition.} We define the input space of images as $\mathcal{X}$, where each image $X \in \mathcal{X}$ is denoted as $X \in \mathbb{R}^{H \times W \times 3}$, $H$ and $W$ being the height and width. The output label space $\mathcal{Y}$ is formed by labels belonging to $K$ semantic categories, such that the one-hot segmentation map can be denoted as $Y \in \mathbb{R}^{H \times W \times K}$. In \task, we are given a source domain dataset with annotations $\mathcal{D}_\mathrm{S} = \{(X_i^\mathrm{S}, Y_i^\mathrm{S})\}^{n_\mathrm{S}}_{i=1}$ and an unlabelled target domain dataset $\mathcal{D}_\mathrm{T} = \{X_i^\mathrm{T}\}^{n_\mathrm{T}}_{i=1}$, such that $p(\mathcal{X}^\mathrm{S}) \neq p(\mathcal{X}^\mathrm{T})$. The objective of \task is to learn a mapping function $f \colon \mathcal{X} \to \mathcal{Y}$ that can correctly predict unlabelled target samples. The function $f = f_c \circ f_b$ is modeled by a neural network, such that it is a composition of the backbone feature extractor $f_b$ and the segmentation decoder $f_c$, which is parameterized by $\theta = \{\theta_b, \theta_c\}$.

\vspace{1mm}

\noindent\textbf{Self-training.} It has been shown in the DASS literature~\cite{daformer, dacs, sepico} that self-training (ST) is an effective technique to reduce the domain gap, which we adopt as a baseline. In details, the ST approach uses the student-teacher (or mean teacher~\cite{mt}) model, where the teacher network $\tilde{f}$ provides \textit{one-hot} pseudo-labels $\hat{Y}$ on-the-fly for the unlabelled target samples to train the student network $f$:

\begin{equation}
    \label{eqn:pseudo-label-mt}
    \hat{Y}^\mathrm{T}({j}) = \mathbf{e}(\arg\max_{c \in \mathcal{Y}} \{\tilde{p}^c_j \colon c \in \mathcal{Y}\})
\end{equation}
where $\tilde{p}^{c}_{j} = \tilde{f}(X^\mathrm{T}_j)$ is the target network prediction probability at pixel $j$ for class $c$, and $\mathbf{e}(\cdot)$ is the one-hot operator. The pseudo-labelled target data is then used to train the student network using a standard cross-entropy (CE) loss:

\begin{equation}
    \label{eqn:ce-loss}
    \mathcal{L}_\text{CE} = - \frac{1}{HW} \frac{1}{K}\sum_{j=1}^{H\times W}\sum_{c=1}^K \hat{Y}_c^{\mathrm{T}}(j) \log p^{c}_{j}
\end{equation}
where $p^c_j = f(X^\mathrm{T}_j)$ is the student network prediction probability for the $j^\textsuperscript{th}$ pixel to be belonging to class $c$ and $\hat{Y}^\mathrm{T}(j)$ is the corresponding one-hot pseudo-label obtained using Eq.~(\ref{eqn:pseudo-label-mt}). Besides the ST objective on the target data, we also optimize the Eq.~(\ref{eqn:ce-loss}) for the annotated source data using the ground-truth labels $Y^\mathrm{S}$.

In ST, the parameters of the teacher network $\tilde{\theta}$ are obtained by taking an exponential moving average (EMA) of the student network parameters $\theta$ at every iteration $t$ as:

\begin{equation}
    \label{eqn:ema-update}
    \tilde{\theta}_{t+1} \leftarrow \beta\tilde{\theta}_t + (1-\beta)\theta_t
\end{equation}
where $\beta$ is a momentum update hyperparameter, which is in general set to 0.999. Note that we optimize the Eq.~(\ref{eqn:ce-loss}) on mixed target images that are obtained using the ClassMix~\cite{dacs} augmentation, instead of the real target images, in order to avoid ST with noisy pseudo-labels.

\subsection{Contrastive Learning Framework for DASS}
\label{sec:contrast-framework}

In this work we propose \method, a contrastive learning framework that enables both \textit{feature-level} and \textit{pixel-level} alignment using the contrastive formulation InfoNCE~\cite{gutmann2010noise}. Our choice of using the contrastive formulation is motivated by the fact that InfoNCE has proven to be beneficial for learning compact pixel embedding space for supervised segmentation~\cite{wang2021exploring} and accurate style transfer~\cite{cut}. We argue that compact pixel representation and accurate style transfer are two key ingredients to attain feature-level and pixel-level alignment, respectively. Given, the feature-level and pixel-level alignment are proven to be two essential ingredients for an effective DASS method (\eg, CyCADA~\cite{cycada}), we revisit the two learning fronts by employing contrastive learning and bring them into an unique framework. Below we elaborate in detail the two alignment techniques.

\subsubsection{Contrastive feature-level alignment}
\label{sec:feat-align}
The feature-level alignment is carried out in the latent feature space of the network by adopting the \textit{pixel-to-prototype} contrast, with the aim of enforcing the pixels of the same semantic category across domains to be close in the embedding space. The driving force behind adopting such a formulation is that the ST training objective only takes into account the `local' context around a given pixel and completely ignores the `global' context from other samples in the dataset and beyond. Moreover, being in the unsupervised scenario, \textit{pixel-to-pixel} contrast~\cite{wang2021exploring} with noisy pseudo-labels of Eq.~(\ref{eqn:pseudo-label-mt}) will lead to reduced performance.

\begin{figure}
    \centering
    \includegraphics[width=\columnwidth]{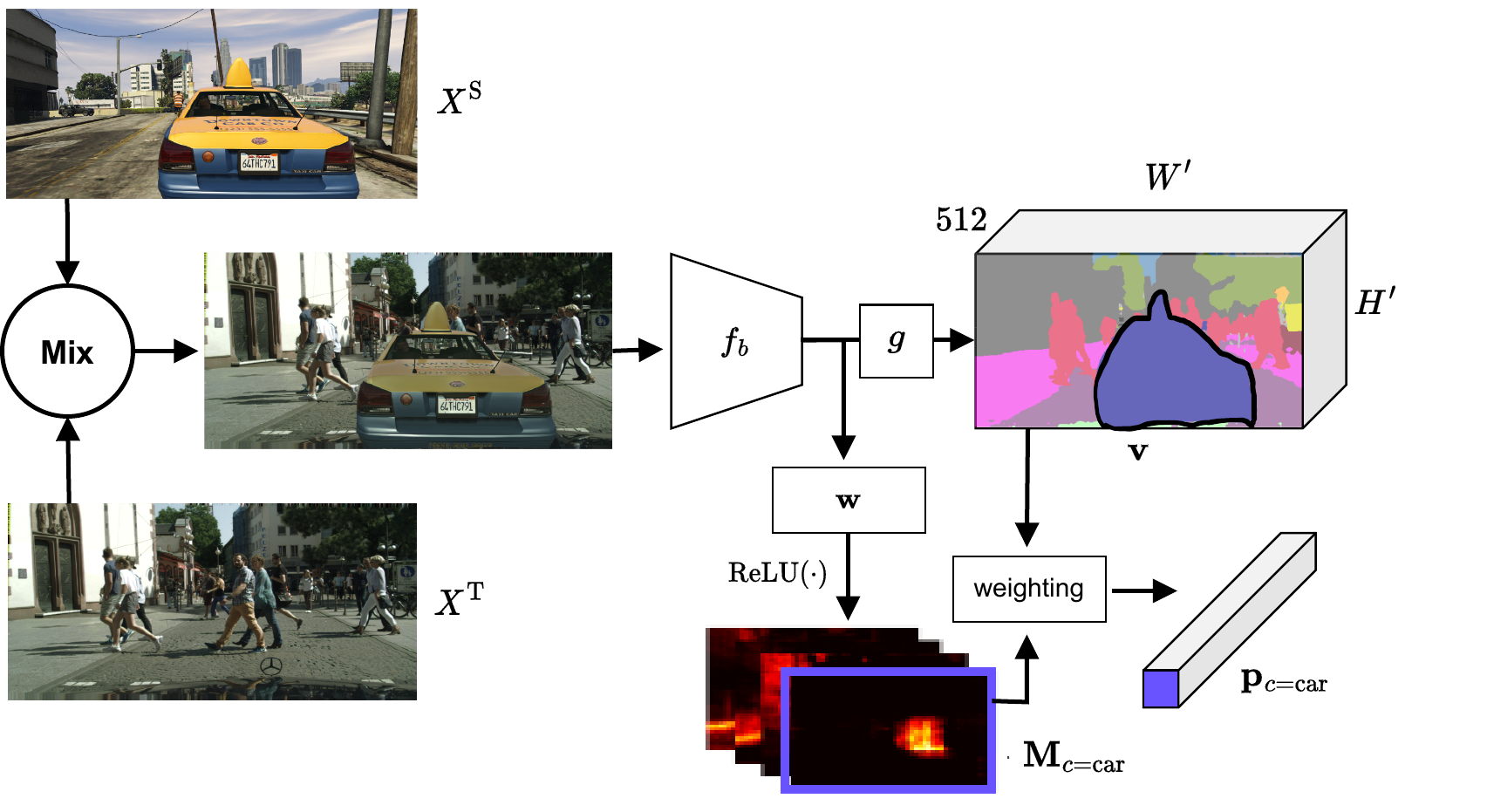}
    \caption{Overview of weighted prototype estimation. The features $\mathbf{v}$ of the mixed image at each spatial location is weighted with the CAM $\mathbf{M}_c$ of each class to obtain a prototype $\mathbf{p}_c$. $\mathbf{v}$ is overlayed with the segmentation map for ease of visualization}
    \label{fig:proto-estimation}
    \vspace{-3mm}
\end{figure}

Guided by these insights we first construct semantic prototypes (or class centroids), one per class, and use these prototypes to pull pixels of the positive class together and at the same time push away pixels from negative classes. Constructing and updating the prototypes is a design choice in itself, and are mainly updated using arithmetic mean~\cite{proca, sepico} or the exponential moving average~\cite{proca, proda}. However, the direct update of the prototype using all the features is unlikely to be applied to the target domain due to the possible erroneous and noisy pseudo-labels, which may lead to inaccurate prototype estimation. Therefore, we employ a weighted prototype estimation method based on the class activation map (CAM)~\cite{zhou2016learning,wsss}.

\vspace{1mm}

\noindent\textbf{Weighted prototype estimation.} The CAM highlights the most discriminative pixel locations in an image that a network looks at for predicting a given class. We employ the CAM in DASS in order to estimate the prototypes. The idea is to compute a weighted average of the embeddings from pixel locations that are maximally activated by CAM for a given class. This results into a prototype that most likely represents the class under consideration. Although, CAM may fail to highlight precise boundary regions of objects, it does not impact our algorithm as the boundary pixel features do not define the canonical representation of objects.

Concretely, features $\mathbf{f} = f_b(X) \in \mathbb{R}^{H'\times W'\times D}$ are obtained from the feature extractor, followed by applying Global Average Pooling (GAP) to collapse the spatial dimensions; where $H'$, $W'$ and $D$ represent the dimensions of the intermediate feature maps. To get the CAM for a particular class $c$, a fully connected layer, having parameters $\mathbf{w} \in \mathbb{R}^{K \times D}$, is learned that outputs a score for each class $c$ as:
\begin{equation}
    \label{eqn:cam-score}
    s_c = \frac{1}{H'W'}\displaystyle \sum^{D}_{d=1} \mathbf{w}_{c, d} \displaystyle \sum^{H' \times W'}_{j=1} \mathbf{f}_{d, j}
\end{equation}
The CAM, denoted as $\mathbf{M}_c$, is then computed for each class as:
\begin{equation}
    \label{eqn:cam}
    \mathbf{M}_c = \textrm{ReLU}\big(\displaystyle \sum^{D}_{d=1} \mathbf{w}_{c,d} \mathbf{f}_{d, :}\big)
\end{equation}
where the $\textrm{ReLU}(\cdot)$ is applied to ignore all negative values. Note that one CAM $\mathbf{M}_c$ is obtained per image.

As shown in Fig.~\ref{fig:proto-estimation}, the prototypes for each class $c$ are then estimated using the just computed $\mathbf{M}_c$ and the projected intermediate features of the images in a mini-batch as:
\begin{equation}
    \label{eqn:cam-proto}
    \mathbf{p}'_c = \frac{\sum_{j \in \mathcal{N}_c} \mathbf{M}_{c,j} \mathbf{v}_{j}}{\sum_{j' \in \mathcal{N}_c} \mathbf{M}_{c,j'}}
\end{equation}
where $\mathcal{N}_c$ denotes the pixel locations in the entire dataset that correspond to the top-\textit{n} highest CAM activation values for class $c$, and $\mathbf{v}_j = g(\mathbf{f}_j) \in \mathbb{R}^{512}$ are the projected features obtained using a non-linear projection head $g(\cdot)$. Note that the prototypes are computed using the teacher network.

\begin{figure*}[!t]
    \centering
    \includegraphics[width=0.85\textwidth]{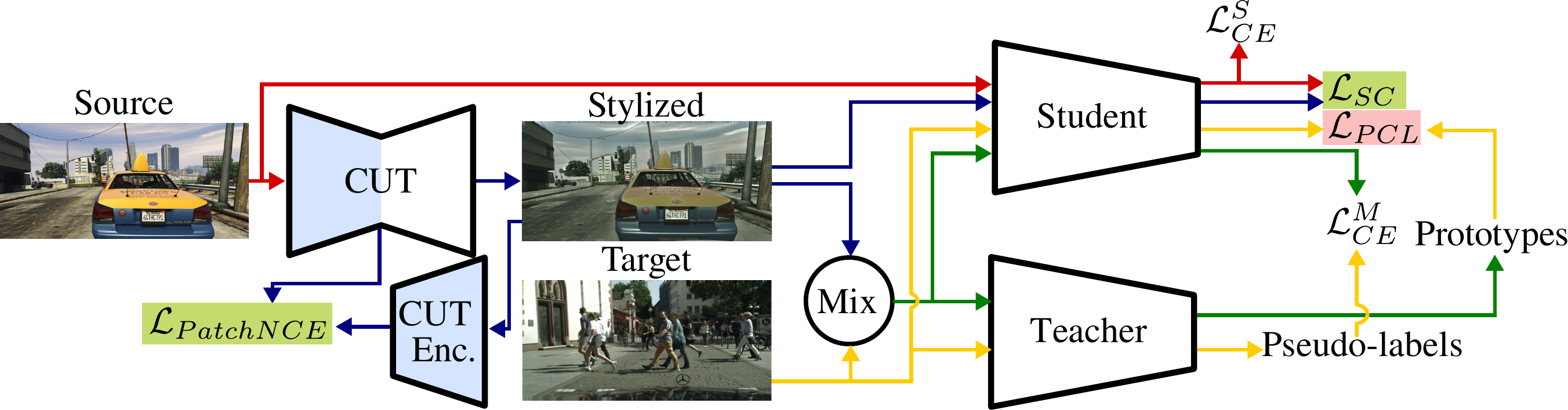}
    \caption{Overview of the proposed \textbf{\method}, which is a mean-teacher framework that unifies {\setlength{\fboxsep}{0pt} \colorbox{protoCol}{feature alignment}} with {\setlength{\fboxsep}{0pt} \colorbox{SpringGreen}{pixel alignment}}. The feature alignment exploits the prototypical contrastive loss ($\mathcal{L}_\text{PCL}$) with mixed-prototypes to align the domains in the feature space. The pixel-alignment exploits contrastive learning ($\mathcal{L}_\text{PatchNCE}$ and $\mathcal{L}_\text{SC}$) to carry out style transfer, aligning the two domains in the pixel space}
    \label{fig:pipeline}
\end{figure*}

The prototypes are then updated in an online manner using the EMA of the prototypes from each mini-batch and the past ones as:
\begin{equation}
    \label{eqn:cam-proto-update}
    \mathbf{p}_{c} \leftarrow \gamma \mathbf{p}_c + (1-\gamma) \mathbf{p}_c'
\end{equation}
where $\mathbf{p}_c$ is the CAM-based prototype of class $c$ for the whole dataset and $\gamma$ being the momentum update hyperparameter. Importantly, instead of estimating prototypes only on the source domain, we apply the update on the mixed image of the source and target domain. Such an update setting can facilitate the reduction of the domain gap by applying the subsequent contrastive loss, which sets us apart from other works using prototypes~\cite{wsss,marsden2022contrastive,proca}.

\paragraph{Prototypical contrastive alignment.} In order to bring closer the representation of the pixels that belong to the same semantic category, we adopt the prototypical contrastive loss (PCL)~\cite{li2021prototypical}, originally designed for image-level representation learning. In details, the PCL ensures that the representation of a sample (\textit{pixel} in our case) is more similar to its corresponding prototype than other unrelated ones. Given, our prototypes are computed using the mixed images, they can be seen as a `bridge' between the source and target domain. Thus, minimizing the PCL translates to aligning the two domains into a shared embedding space. Our proposed method differs from the previous DASS works~\cite{sepico, proca}, which also adopt the PCL, in the manner in which the prototypes are computed.

Given a projected feature $\mathbf{v}_j$ extracted by the student $g \circ f_b$ corresponding to a pixel $X_j$ and the estimated prototypes $\mathcal{P} = \{\mathbf{p}_c\}_{c=1}^{K}$, the pixel-to-prototype likelihood for pixel $j$ is given as:
\begin{equation}
    \label{eqn:pcl-likelihood}
    p_{j, c} = \frac{\exp(\mathbf{v}_j \cdot \mathbf{p}_c / T)}{\sum_{k=1}^K \exp(\mathbf{v}_j \cdot \mathbf{p}_k / T)}
\end{equation}
where $\mathbf{p}_c$ is the prototype belonging to the same class as pixel $X_j$ and $T$ is the temperature. To attract the feature to prototype of class $c$ and repel it from others, the following loss is adopted:
\begin{equation}
    \label{eqn:pcl}
    \mathcal{L}_\text{PCL} = - \frac{1}{N} \sum_{j=1}^{N} Y(j, c) \log p_{j, c}
\end{equation}
where $N$ is the number of randomly sampled features to avoid performance degradation caused by erroneous labels, and $Y$ is the one-hot ground truth mask for the source domain features and pseudo-labels for the target domain features.

\vspace{-5mm}
\subsubsection{Contrastive pixel-level alignment}
\label{sec:pixel-align}

In order to further mitigate the domain-shift, \method enables pixel-level alignment by generating target-\textit{like} source samples that have the same content as the source images but appear to be drawn from the target domain. As shown in the previous generative DASS approaches~\cite{cycada,sankaranarayanan2018generate,russo2018source}, pixel-level alignment can sometimes outperform feature-level alignment methods.

A commonality among these generative DASS approaches is the use of cycle-consistency loss~\cite{zhu2017unpaired} that is susceptible to two major issues: (\textbf{i}) the generator encodes noise (or high frequency signal) during the forward translation, which is then utilized as a \textit{shortcut} during the reverse translation to reconstruct the original image, and (\textbf{ii}) the generator can accurately translate images which adhere to the target domain statistics but dramatically changing the source content. To overcome the drawbacks of bi-directional translation, one-sided unpaired image translation have been proposed~\cite{benaim2017one,fu2019geometry,roy2021trigan,cut}. In this work we adopt CUT~\cite{cut}, an unpaired image translation (or \textit{stylization)} method that exploits the contrastive learning to associate corresponding \textit{patches} in the two domains to be similar.

Concretely, as shown in Fig.~\ref{fig:teaser} we employ the patch-based InfoNCE loss~\cite{cut}, where given the projected features of an anchor patch in the stylized image $\hat{\mathbf{z}}^{\mathrm{S} \to \mathrm{T}}$, the corresponding \textit{positive} patch in the source image $\mathbf{z}^\mathrm{S}_{+}$, and a set of \textit{negative} patches from other locations in the source image $\{\mathbf{z}^{\mathrm{S}}_{-}\}^{n}_{j=1}$ (see the Supplement for details), is given as:

\vspace{-3mm}
\begin{flalign}
    \label{eqn:cut-loss}
    \mathcal{L}_\text{PatchNCE} = - \mathbb{E}_{\hat{\mathbf{z}}^{\mathrm{S} \to  \mathrm{T}}} \log \frac{\exp( \hat{\mathbf{z}}^{\mathrm{S} \to \mathrm{T}} \cdot \mathbf{z}^\mathrm{S}_{+} / T)}{\Big[\splitfrac{\exp( \hat{\mathbf{z}}^{\mathrm{S} \to \mathrm{T}} \cdot \mathbf{z}^\mathrm{S}_{+} / T)}{+ \sum^{n}_{j=1} \exp(\hat{\mathbf{z}}^{\mathrm{S} \to \mathrm{T}} \cdot \mathbf{z}^{\mathrm{S}}_{-, j} / T)}\Big]}
\end{flalign}

To further alleviate the problem of semantic inconsistency in the translated images we propose a joint training of the stylization and the segmentation module. It follows a \textit{virtuous cycle}: better segmentation model leads to high quality image-translation, and better quality domain translation leads to improved segmentation under domain-shift. In details, we additionally propose to use a prototypical semantic consistency loss that uses the prototypes (detailed in Sec.~\ref{sec:feat-align}) to ensure that the target-\textit{like} images do not hallucinate incorrect content during translation, which are not present in the source:
\begin{equation}
    \label{eqn:sem-cons-loss}
    \mathcal{L}_\text{SC} = \frac{1}{H'W'K}\sum_{j=1}^{H'\times W'}\sum_{c=1}^K\|\mathbf{v}^{\mathrm{S}}_j\cdot \mathbf{p}_c - \phi(\hat{\mathbf{v}}^{\mathrm{S} \to \mathrm{T}}_j) \cdot \mathbf{p}_c \|^2
\end{equation}
where $\phi(\cdot)$ is a learnable affine transformation applied on the features from the stylized image to allow gaps between two images, for instance color, lightness and texture.

We further use the prototypes in Eq.~(\ref{eqn:sem-cons-loss}), which models each category, to classify the features extracted by the student's backbone. The semantic consistency loss thus ensures that each pixel corresponds to an identical class after the translation. Note that the gradients from this loss are not used for the optimization of the segmentation network.

\begin{table*}[h]
\centering
\small
\caption{Comparison results on \textbf{GTA5 $\rightarrow$ Cityscapes}. Methods based on \colorbox{SpringGreen}{pixel alignment} are highlighted with colors. $^\dag$ indicate methods trained at higher resolution. The best performance are in \textbf{bold} and the best performance in low resolution setting are marked with \underline{underline}}
\label{tab:gtacs}
\resizebox{\textwidth}{!}{%
\begin{tabular}{@{}c|ccccccccccccccccccc|c@{}}
\toprule
 Methods & \rotatebox{90}{road} & \rotatebox{90}{sideway} & \rotatebox{90}{building} & \rotatebox{90}{wall} & \rotatebox{90}{fence} & \rotatebox{90}{pole} & \rotatebox{90}{light} & \rotatebox{90}{sign} & \rotatebox{90}{vegetation} & \rotatebox{90}{terrace} & \rotatebox{90}{sky} & \rotatebox{90}{person} & \rotatebox{90}{rider} & \rotatebox{90}{car} & \rotatebox{90}{truck} & \rotatebox{90}{bus} & \rotatebox{90}{train} & \rotatebox{90}{motor} & \rotatebox{90}{bike} & mIoU \\ \midrule
\colorbox{SpringGreen}{Cycada} \cite{cycada}& 86.7&35.6&80.1&19.8&17.5&38.0&39.9&41.5&82.7&27.9&73.6&64.9&19.0&65.0&12.0&28.6&4.5&31.1&42.0&42.7 \\
\colorbox{SpringGreen}{Li \etal} \cite{bidir} & 91.0&44.7&84.2&34.6&27.6&30.2&36.0&36.0&85.0&43.6&83.0&58.6&31.6&83.3&35.3&49.7&3.3&28.8&35.6&48.5 \\
\colorbox{SpringGreen}{FDA-MBT} \cite{fda} & 92.5&53.3&82.4&26.5&27.6&36.4&40.6&38.9&82.3&39.8&78.0&62.6&34.4&84.9&34.1&53.1&16.9&27.7&46.4&50.5 \\
DACS \cite{dacs} & 89.9&39.7&87.9&30.7&39.5&38.5&46.4&52.8&88.0&44.0&88.8&67.2&35.8&84.5&45.7&50.2&0.0&27.3&34.0&52.1  \\
CPSL \cite{li2022class}&91.7&52.9&83.6&43.0&32.3&43.7&51.3&42.8&85.4&37.6&81.1&69.5&30.0&88.1&44.1&59.9&24.9&47.2&48.4&55.7\\
\colorbox{SpringGreen}{Ma \etal} \cite{gpa} & 92.5&58.3&86.5&27.4&28.8&38.1&46.7&42.5&85.4&38.4&91.8&66.4&37.0&87.8&40.7&52.4&\textbf{\underline{44.6}}&41.7&59.0&56.1 \\
ProCA \cite{proca} &91.9&48.4&87.3&41.5&31.8&41.9&47.9&36.7&86.5&42.3&84.7&68.4&43.1&88.1&39.6&48.8&40.6&43.6&56.9&56.3 \\
DecoupleNet \cite{lai2022decouplenet} &88.5&47.8&87.4&38.3&36.9&44.9&53.8&39.6&88.0&38.7&88.8&70.4&39.4&87.8&31.4&55.0&37.4&47.1&55.9&56.7\\
ProDA \cite{proda} &87.8&56.0&79.7&\textbf{\underline{46.3}}&\underline{44.8}&\underline{45.6}&53.5&53.5&\underline{88.6}&45.2&82.1&70.7&39.2&88.8&45.5&59.4&1.0&48.9&56.4& 57.5 \\ 

SePiCo\cite{sepico} & 95.2 &  67.8        & 88.7         &  41.4    &   38.4    &  43.4    &  55.5     &  \underline{63.2}    &    \underline{88.6}        &    46.4     &  \underline{88.3}   &  \underline{73.1}      &  49.0     & \underline{91.4}    &  \textbf{\underline{63.2}}     &  \underline{60.4}   &  0.0     &   45.2    & 60.0     &  61.0\\ 
\method (Ours) &   \underline{95.7}   &  \underline{69.9}   &   \underline{89.5}       & 34.6     &  42.6     & 40.9     &  \underline{57.5}     &  59.4    &    \underline{88.6}   &  \textbf{\underline{49.0}}  &  88.2   &     72.8  &  \textbf{\underline{53.4}}     & 90.1    &  61.8     &  54.9   &   13.9    &   \textbf{\underline{50.2}}    &  \underline{63.4}    & \underline{62.2} \\ \midrule
HRDA \cite{hoyer2022hrda}$^\dag$ &   96.2&73.1&\textbf{89.7}&43.2&39.9&47.5&60.0&60.0&\textbf{89.9}&47.1&\textbf{90.2}&75.9&49.0&91.8&61.9&59.3&10.2&47.0&\textbf{65.3}& 63.0 \\
\method (Ours)$^\dag$ &   \textbf{96.5}&\textbf{75.6}&88.9&45.1&\textbf{45.9}&\textbf{50.1}&\textbf{61.2}&\textbf{68.2}&89.4&45.7&86.3&\textbf{76.3}&49.9&\textbf{92.2}&55.1&\textbf{62.8}&16.7&33.8&63.1    & \textbf{63.3} \\\bottomrule
\end{tabular}%
}
\end{table*}

\subsubsection{Training objectives}

The whole pipeline of \method is depicted in the Fig.~\ref{fig:pipeline}. The training will be divided into two phases. In the first, or the \textit{joint} training phase, the overall loss $\mathcal{L}_\text{Joint}$ is given as:
\begin{equation}
\label{eqn:joint-losses}
        \mathcal{L}_\text{Joint} = \underbrace{\mathcal{L}^{\text{S,M}}_\text{CE}}_\text{self-training} 
        + \underbrace{\lambda_\text{PCL}\mathcal{L}^{\text{S,M}}_\text{PCL}}_\text{\colorbox{protoCol}{feature alignment}}  
        +\underbrace{\lambda_\text{style} (\mathcal{L}_\text{PatchNCE} + \mathcal{L}_\text{SC})}_\text{\colorbox{SpringGreen}{pixel alignment}}
\end{equation}
where $\text{S}$ and $\text{M}$ denote that the losses are applied to the source and the mixed images, respectively. $\lambda_\text{PCL}$ and $\lambda_\text{style}$ are used for weighing the corresponding losses.

In order to avoid overfitting the segmentation model to the stylized image features, such as textures, during the joint training with the image-to-image translation model, we propose a second-round training where we train the segmentor from scratch, with the style transfer network kept frozen.

\section{Experiments}
\label{sec:exp}
\subsection{Implementation Details}

\noindent\textbf{Datasets.} We follow the experimental protocols adopted in the previous \task works~\cite{dacs, proca, proda, sepico}. For the source domain, we use the synthetic GTA dataset~\cite{gta} containing $24,966$ synthetic images of resolution $1914\times 1052$ and the SYNTHIA dataset~\cite{synthia} with $9400$ synthetic images of resolution $1280\times 760$. As target domain we use the Cityscapes dataset~\cite{cs} which contains $2975$ training and $500$ test images of resolution $2048\times 1024$. In the low-resolution setting, following \cite{sepico}, images are resized to $1280\times 640$ for Cityscapes dataset and to $1280\times 720$ for GTA dataset before randomly cropping to $640\times 640$. For fair comparison with HRDA~\cite{hoyer2022hrda}, we also perform experiments in full resolution and use $1024\times1024$ crops for training.

\noindent\textbf{Training.} As in \cite{dacs, proca, proda, sepico}, we adopt DeepLab-V2~\cite{deeplab} with ResNet-101~\cite{resnet} as backbone.  We use the AdamW optimizer \cite{adamw} with the initial learning rate set to $6\times 10^{-5}$ and weight decay of $0.01$. We adopt the warm-up policy as well as the rare class sampling proposed by~\cite{daformer}. Following \cite{dacs}, we apply the color jittering, Gaussian blurring and ClassMix~\cite{dacs} on the mixed images.

\subsection{Comparison with the State-of-the-Art}
\label{sec: cpsota}

\begin{table*}[h]
\centering
\small
\caption{Comparison results on \textbf{SYNTHIA $\rightarrow$ Cityscapes}. Methods based on \colorbox{SpringGreen}{pixel alignment} are highlighted with colors} 
\label{tab:synthiacs}
\resizebox{\textwidth}{!}{%
\begin{tabular}{@{}c|cccccccccccccccc|cc@{}}
\toprule
 & \rotatebox{90}{road} & \rotatebox{90}{sideway} & \rotatebox{90}{building} & \rotatebox{90}{wall*} & \rotatebox{90}{fence*} & \rotatebox{90}{pole*} & \rotatebox{90}{light} & \rotatebox{90}{sign} & \rotatebox{90}{vegetation} & \rotatebox{90}{sky} & \rotatebox{90}{person} & \rotatebox{90}{rider} & \rotatebox{90}{car} & \rotatebox{90}{bus} & \rotatebox{90}{motor} & \rotatebox{90}{bike} & mIoU & mIoU* \\ \midrule
 \colorbox{SpringGreen}{Li \etal} \cite{bidir} & 86.0&46.7&80.3&-&-&-&14.1&11.6&79.2&81.3&54.1&27.9&73.7&42.2&25.7&45.3&-&51.4 \\
 \colorbox{SpringGreen}{FDA-MBT} \cite{fda} & 79.3&35.0&73.2&-&-&-&19.9&24.0&61.7&82.6&61.4&31.1&83.9&40.8&38.4&51.1& - &52.5 \\
DACS \cite{dacs} & 80.6&25.1&81.9&21.5&2.9&37.2&22.7&24.0&83.7&\textbf{90.8}&67.6&38.3&83.0&38.9&28.5&47.6&48.3&54.8  \\ 
\colorbox{SpringGreen}{Ma \etal} \cite{gpa} & 75.7&30.0&81.9&11.5&2.5&35.3&18.0&32.7&86.2&90.1&65.1&33.2&83.3&36.5&35.3&54.3&48.2&55.5 \\
ProCA \cite{proca} &\textbf{90.5}&\textbf{52.1}&84.6&29.2&3.3&40.3&37.4&27.3&86.4&85.9&69.8&28.7&88.7&53.7&14.8&54.8&53.0&59.6 \\ 
CPSL \cite{li2022class} &87.3&44.4&83.8&25.0&0.4&42.9&47.5&32.4&86.5&83.3&69.6&29.1&89.4&52.1&42.6&54.1&54.4&61.7\\
ProDA \cite{proda} &87.8&45.7&84.6&\textbf{37.1}&0.6&44.0&54.6&37.0&\textbf{88.1}&84.4&74.2&24.3&88.2&51.1&40.5&45.6&55.5&62.0 \\ 
DecoupleNet \cite{lai2022decouplenet} &77.8&48.6&75.6&32.0&1.9&\textbf{44.4}&52.9&38.5&87.8&88.1&71.1&34.3&88.7&58.8&50.2&61.4&57.0&64.1\\
SePiCo\cite{sepico} & 77.0 &  35.3        & 85.1         &  23.9    &   3.4    &  38.0    &  51.0     & 55.1     &    85.6        &    80.5     &  73.5   &  46.3      &  87.6     & \textbf{69.7}    &  \textbf{50.9}     &  \textbf{66.5}   &  58.1     &   66.5  \\ 
\method (Ours) &   83.8   &  44.6   &   \textbf{86.9}       & 15.4     &  \textbf{3.7}     & 44.3     &  \textbf{56.9}     &  \textbf{55.5}    &    84.9   &  86.2  &  \textbf{73.8}   &     \textbf{46.8}  &  \textbf{90.1}     & 57.1    &  46.0     &  63.2   &   \textbf{58.7}    & \textbf{67.4}   \\ \bottomrule

\end{tabular}%
}
\end{table*}

We compare with recent state-of-the-art methods \cite{cycada, bidir, fda, dacs, li2022class, gpa, proca, proda, lai2022decouplenet, sepico, hoyer2022hrda}, especially those using style transfer \cite{bidir, cycada, gpa, fda} and prototypes \cite{proca, proda, sepico}. On the GTA $\rightarrow$ Cityscapes task, \method is compared separately with HRDA\cite{hoyer2022hrda} since it operates at full resolution.

The quantitative comparison on GTA $\rightarrow$ Cityscapes is reported in \cref{tab:gtacs}. We observe that our approach outperforms prior methods with the mIoU of $62.2\%$.  In particular, our method shows its high capacity on easy to confuse class pairs, such as motor-bike, road-sideway, and person-rider pairs. Moreover, working at higher resolution improves the performance on small objects, especially on challenging classes such as `train', with an overall improvement of 0.3\% over HRDA~\cite{hoyer2022hrda}. As shown in the~\cref{tab:synthiacs} for the SYNTHIA $\rightarrow$ Cityscapes benchmark, the proposed method also obtains state-of-the-art performance. More precisely, we obtain $58.7\%$ and $67.4\%$ mIoU in the 16-category and 13-category evaluation protocols, respectively. In both the benchmarks, we observe that existing methods based on style transfer underperform more recent methods based on feature alignment and self-training. Overall our \method demonstrates that unifying these two orthogonal research directions can lead to state-of-the-art results.

\subsection{Ablation Studies}

\begin{table}[h]
\centering
\small
\caption{Ablation study on the \textbf{GTA5} $\rightarrow$ \textbf{Cityscapes}. PCL denotes the use of prototypical contrastive learning}
\label{tab:abla}
\resizebox{0.9\columnwidth}{!}{%
    \begin{tabular}{l|cccc|cc}
    \toprule
    \multirow{2}{*}{Method} & \multicolumn{3}{c}{Style Transfer} & \multirow{2}{*}{PCL} &  \multirow{2}{*}{mIoU} & \multirow{2}{*}{$\Delta$} \\
    & Offline & Joint & Two-stage & & & \\ 
    \midrule
    
     Baseline (\cref{sec:prel}) & & & & & 57.5 & -\\
     \textbf{A} & \checkmark &  &  & & 57.1 & \textcolor{red}{-0.4} \\
     \textbf{B} (Feature align) & & & & \checkmark & 60.0 & \textcolor{darkgreen}{+2.5} \\
 
     \textbf{C} & \checkmark & & & \checkmark & 57.6 & \textcolor{darkgreen}{+0.1}\\
     \textbf{D} (Pixel align) & & \checkmark & & & 59.0 & \textcolor{darkgreen}{+1.5} \\
     \textbf{E} & & \checkmark & \checkmark & & 59.0 & \textcolor{darkgreen}{+1.5} \\
     \method (Ours) & & \checkmark & \checkmark & \checkmark &  \textbf{62.2} & \textcolor{darkgreen}{\textbf{+4.7}} \\
    
    \bottomrule
\end{tabular}}
\vspace{-3mm}
\end{table}

\noindent\textbf{Evaluation of the proposed pipeline.}
We thoroughly ablate our proposed \method in order to measure the impact of: (\textbf{i}) joint training of the style-transfer and segmentation models, (\textbf{ii}) our two-stage training procedure, and (\textbf{iii}) the introduction of prototypes in our pixel-alignment technique. In these ablations, we start from the self-training baseline which is described in~\cref{sec:prel}.

We report the results of the ablations in~\cref{tab:abla}.
First, when the style transfer network, which aligns domains at pixel-level, is trained separately from the segmentation network (see model \textbf{A} in~\cref{tab:abla}), we observe a performance drop of 0.4\% mIoU w.r.t the baseline. It indicates that stylization is not able to bridge the domain gap. On the contrary, when we perform feature alignment via prototypical contrastive learning but without any stylization, we observe a clear gain of +2.5\% (see model \textbf{B}). Surprisingly, including offline stylization into the pipeline (see model \textbf{C}) is again detrimental (57.6\% vs +60.0\%). This may be explained by inaccurate stylization if not jointly done, which alters the image content and makes the estimated prototypes noisy.

When the stylization is no longer trained offline but jointly with the segmentation model (see model \textbf{D}), we start observing gains in performance ($+1.5\%$ compared to the baseline). Then, our two-stage training without $\mathcal{L}_\text{PCL}$, \ie, model \textbf{E}, does not improve the performance w.r.t model \textbf{D}, which shows the limits of stand-alone pixel-level alignment. However, when we combine joint style transfer and prototypical contrastive in a two stage training fashion, \method achieves the best performance of $62.2\%$, which is $+4.7\%$ higher than the baseline. These ablations justify that both the feature-level and pixel-level alignment contribute constructively by reducing the domain gap: joint training of the style transfer model improves pixel alignment and prototype estimation while the feature alignment loss promotes domain invariant features.

\def\myim#1{\includegraphics[height=18.5mm]{img/#1}}

\begin{figure*}
  \setlength\tabcolsep{1. pt}
\centering
 \begin{tabular}{ccccc}

    Source image & CUT~\cite{cut} & AdaIN \cite{adain} & GPA \cite{gpa} & FDA \cite{fda}
    \\\myim{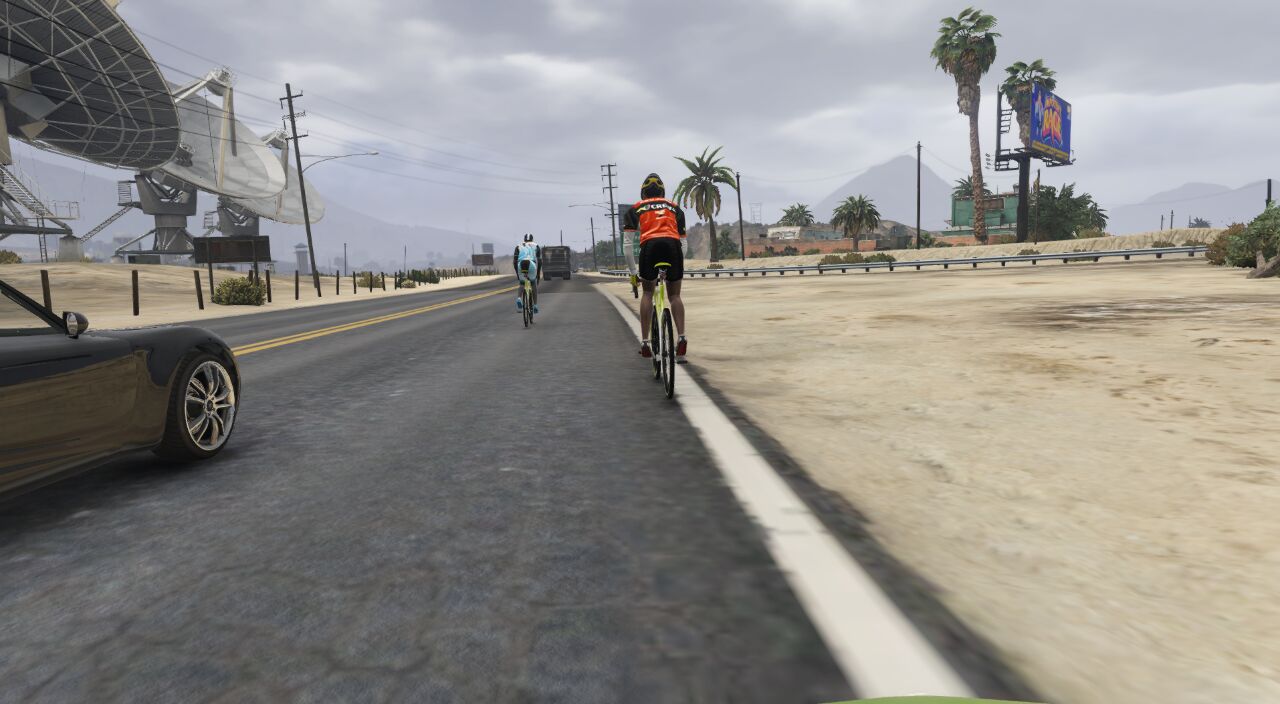} &
    \myim{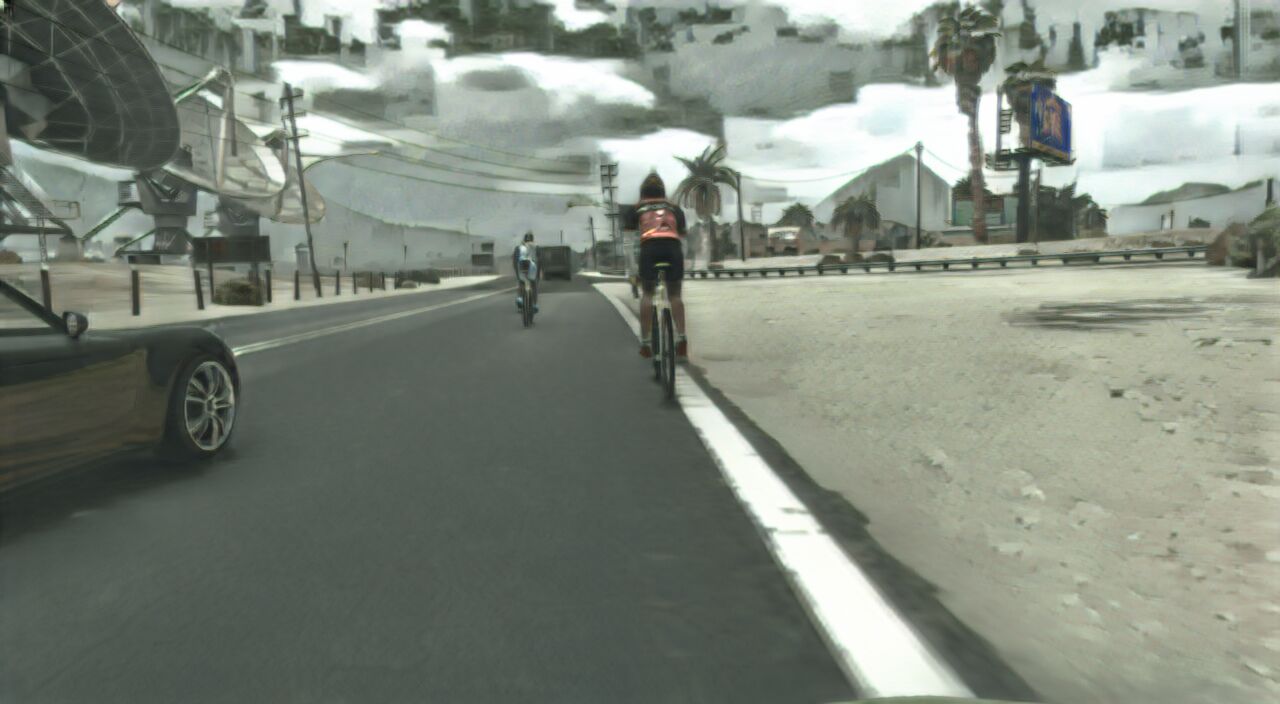} &
    \myim{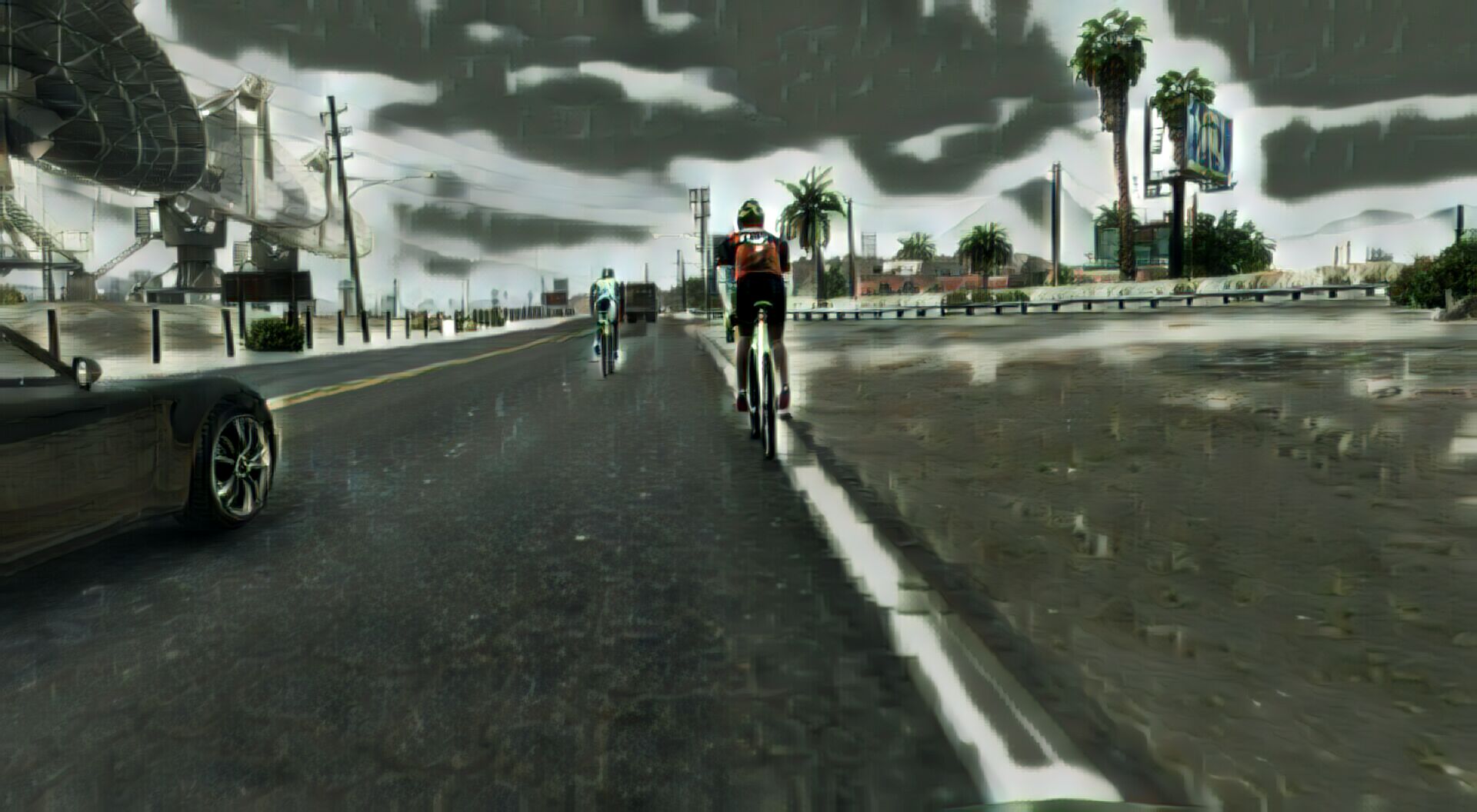}  &
    \myim{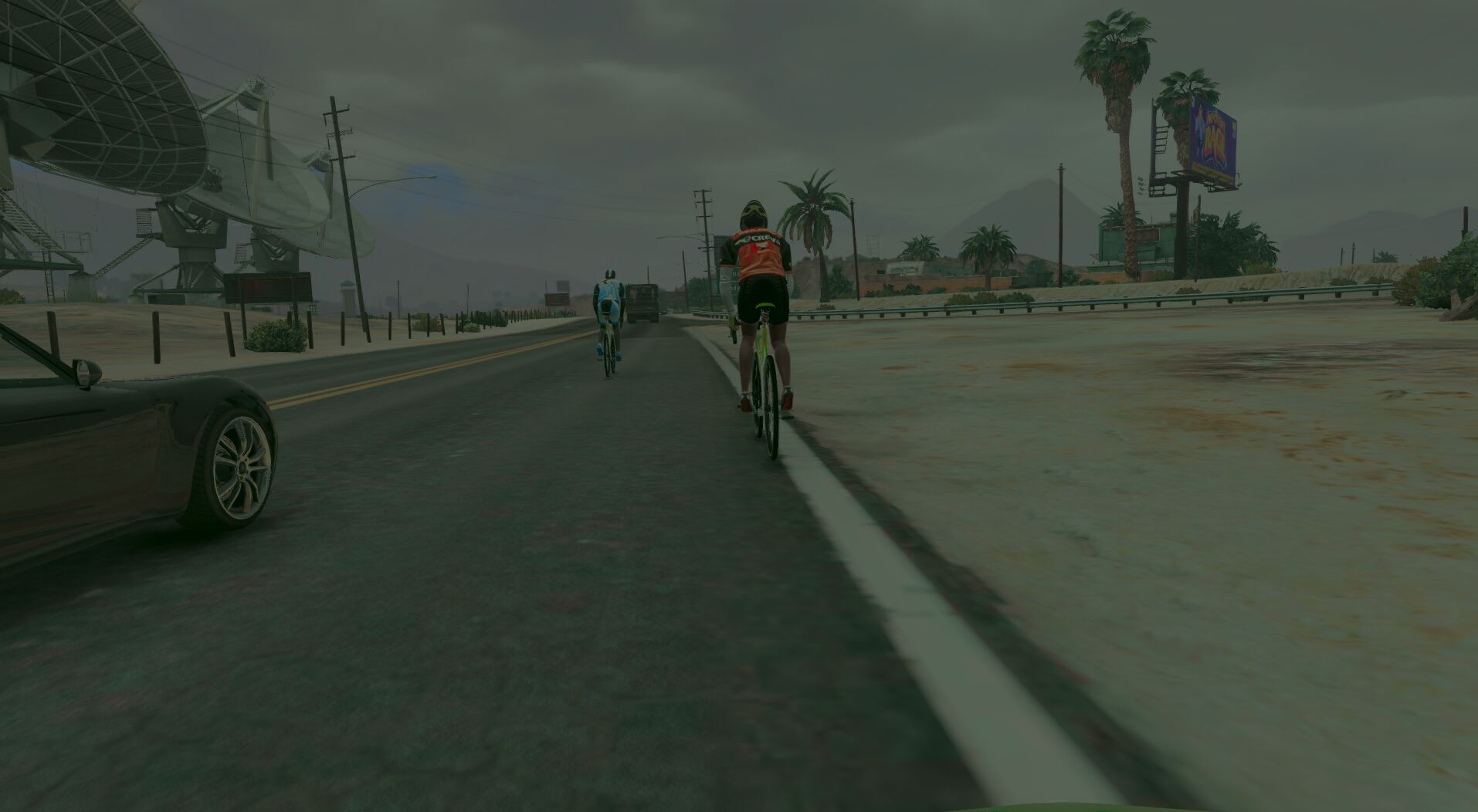}&\myim{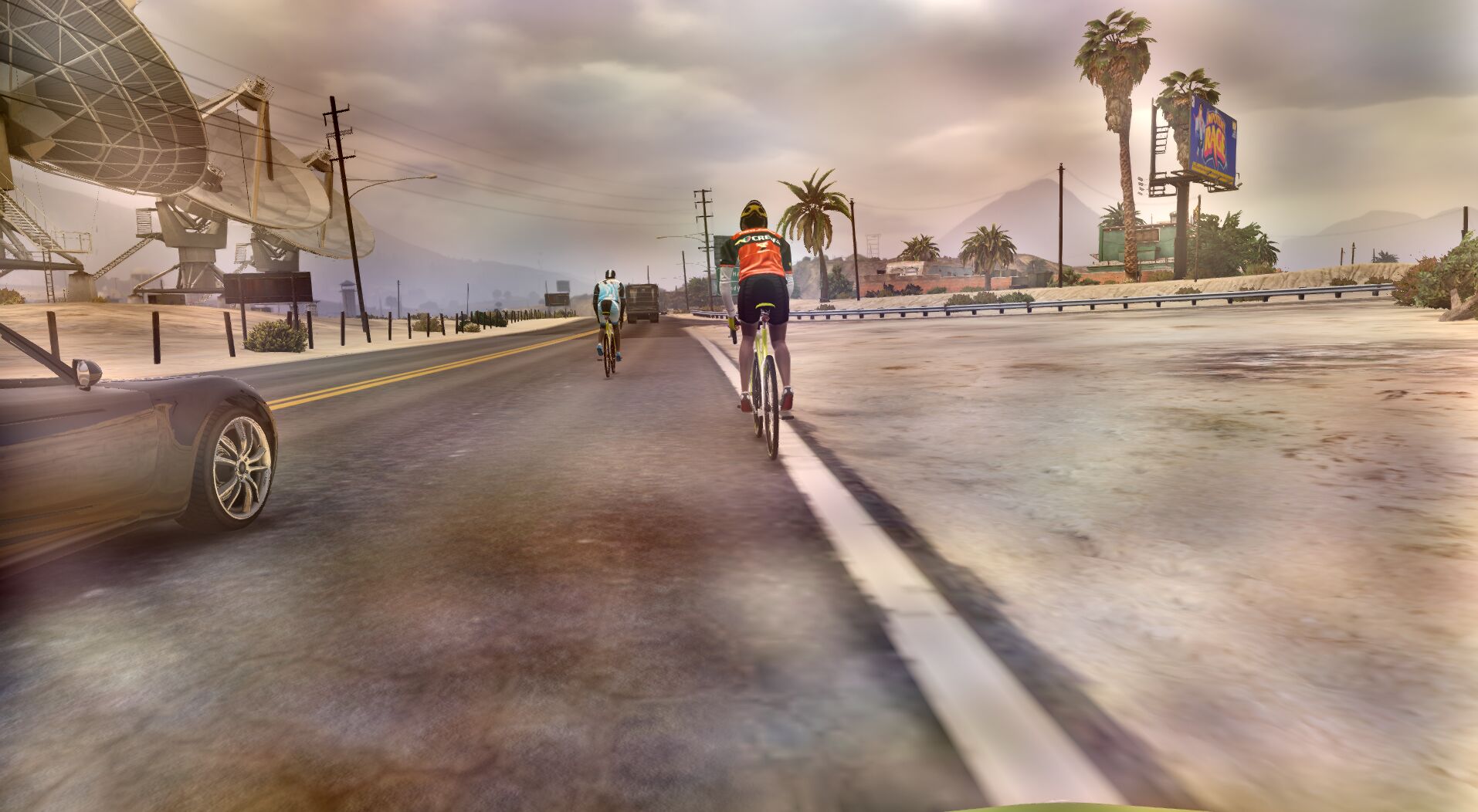} \\

    Target image
    &w/o $\mathcal{L}_{SC}$
   & $\mathcal{L}_{CE}$
      &MSE
    &\method(Ours)\\
    \myim{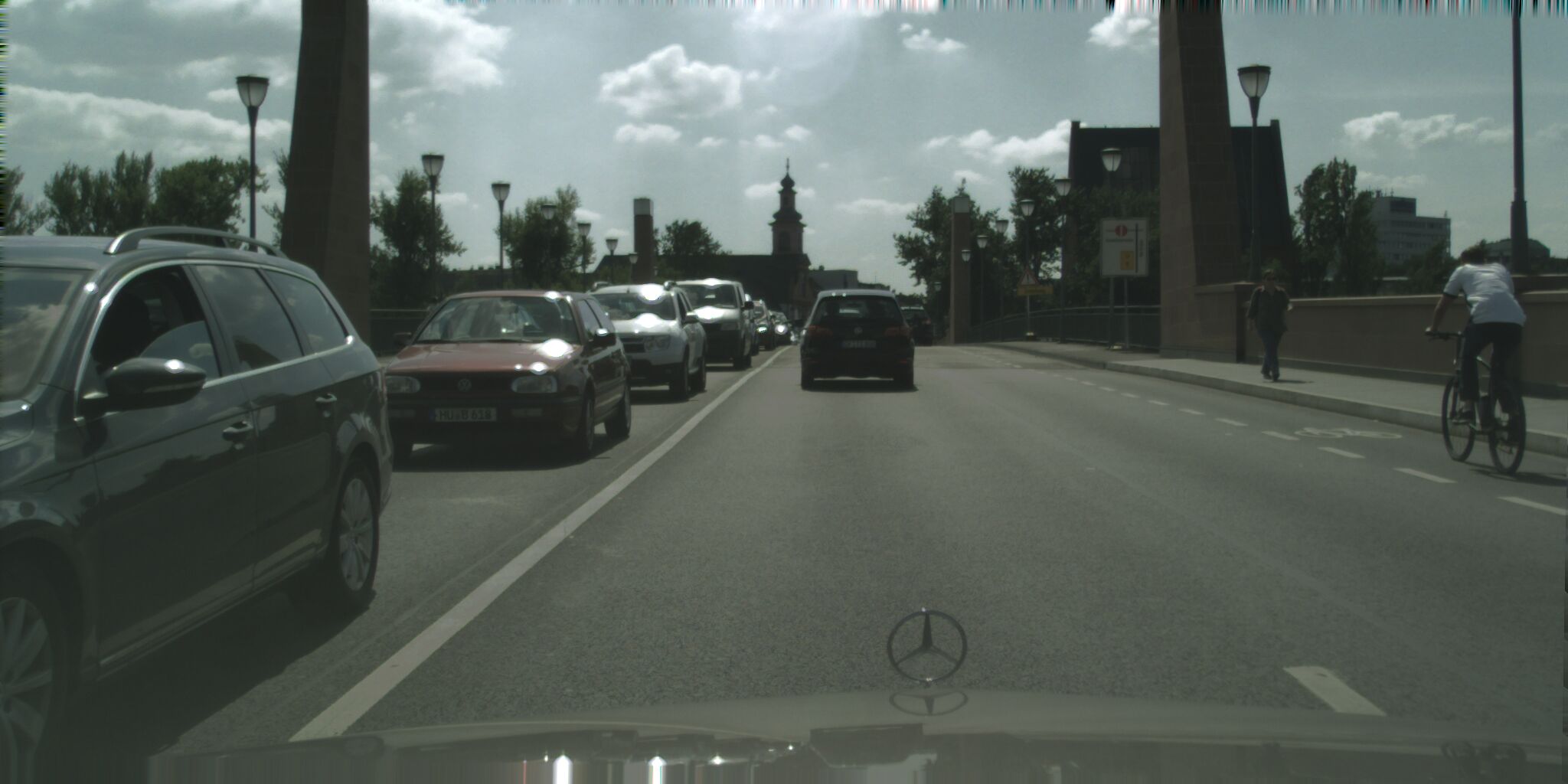}
    &\myim{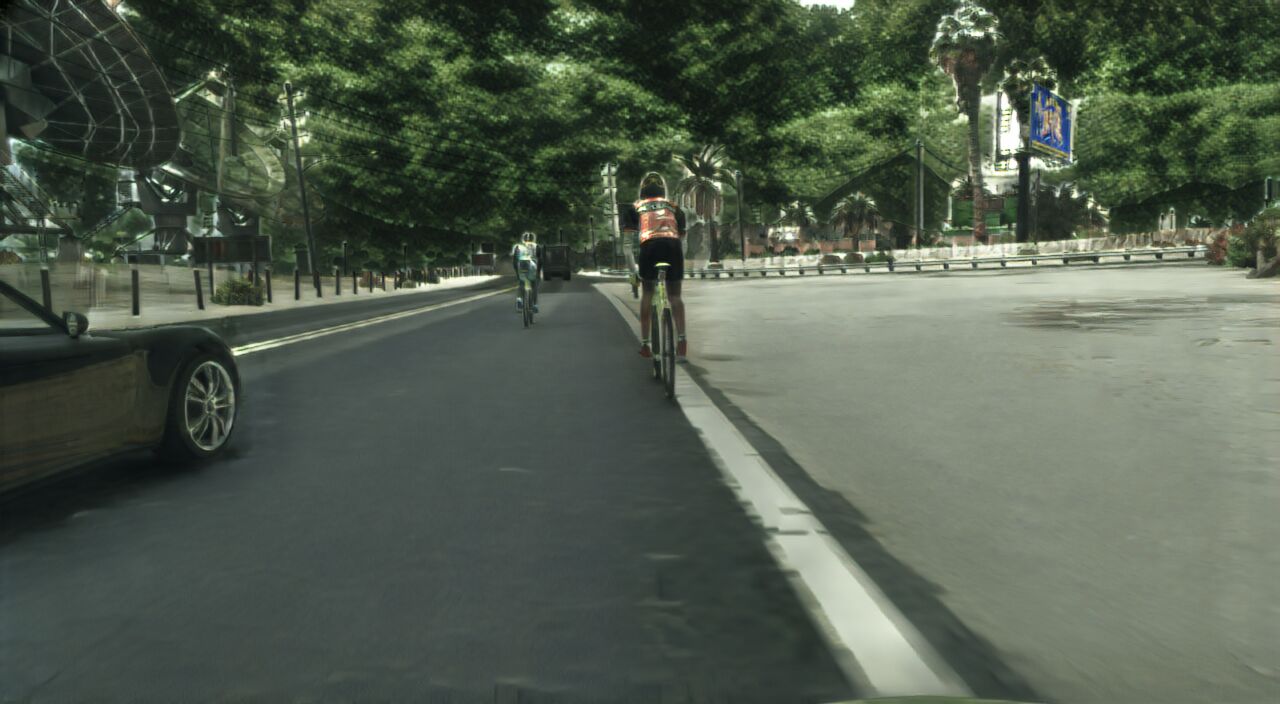}
   & \myim{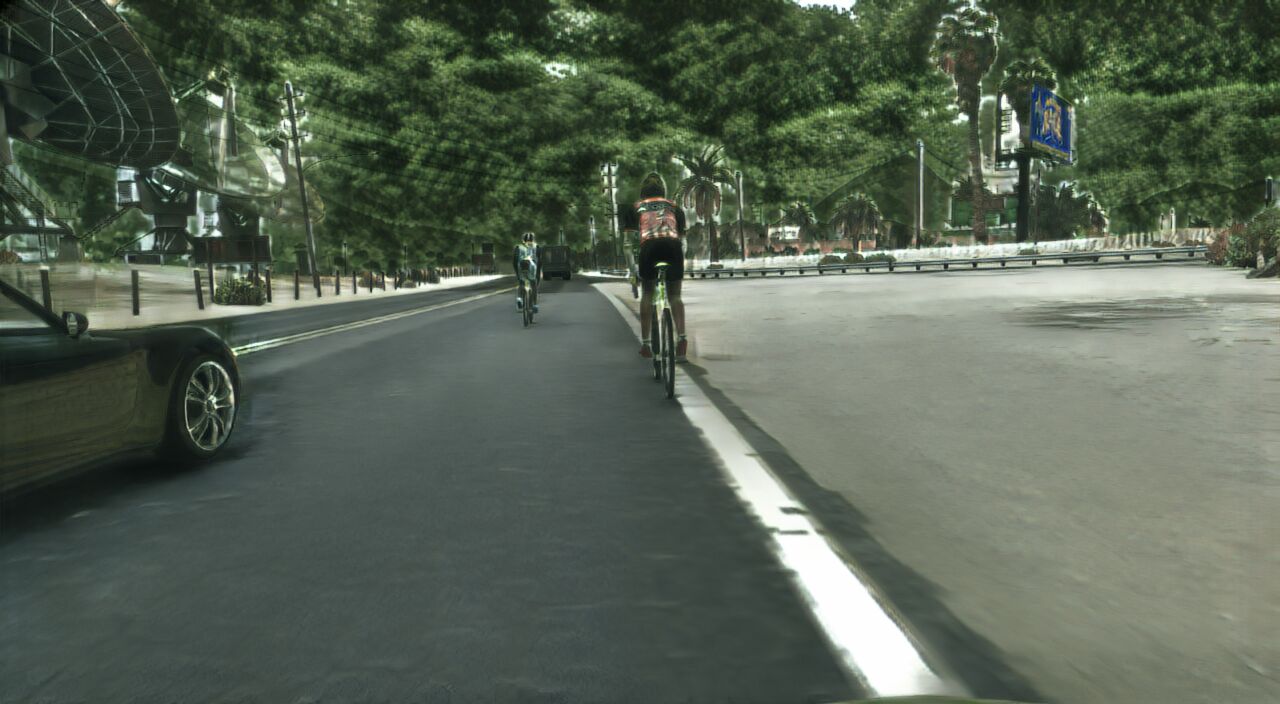}

    &\myim{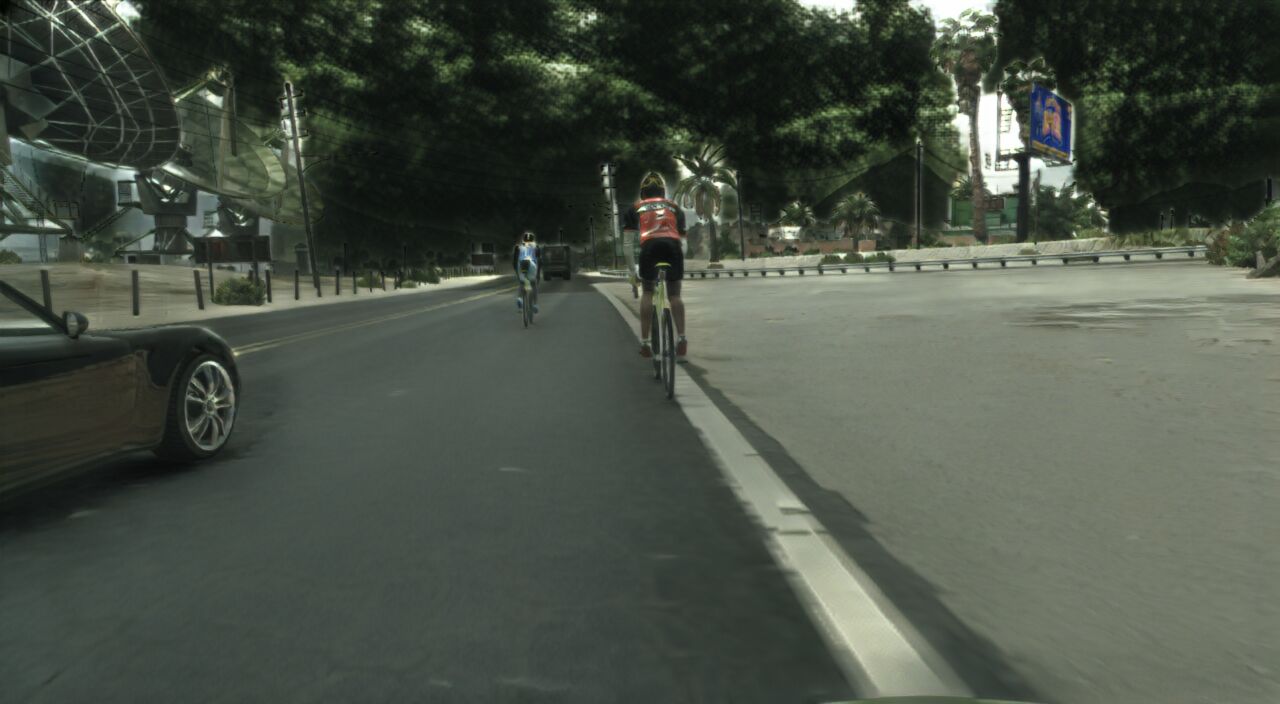}  &
    \myim{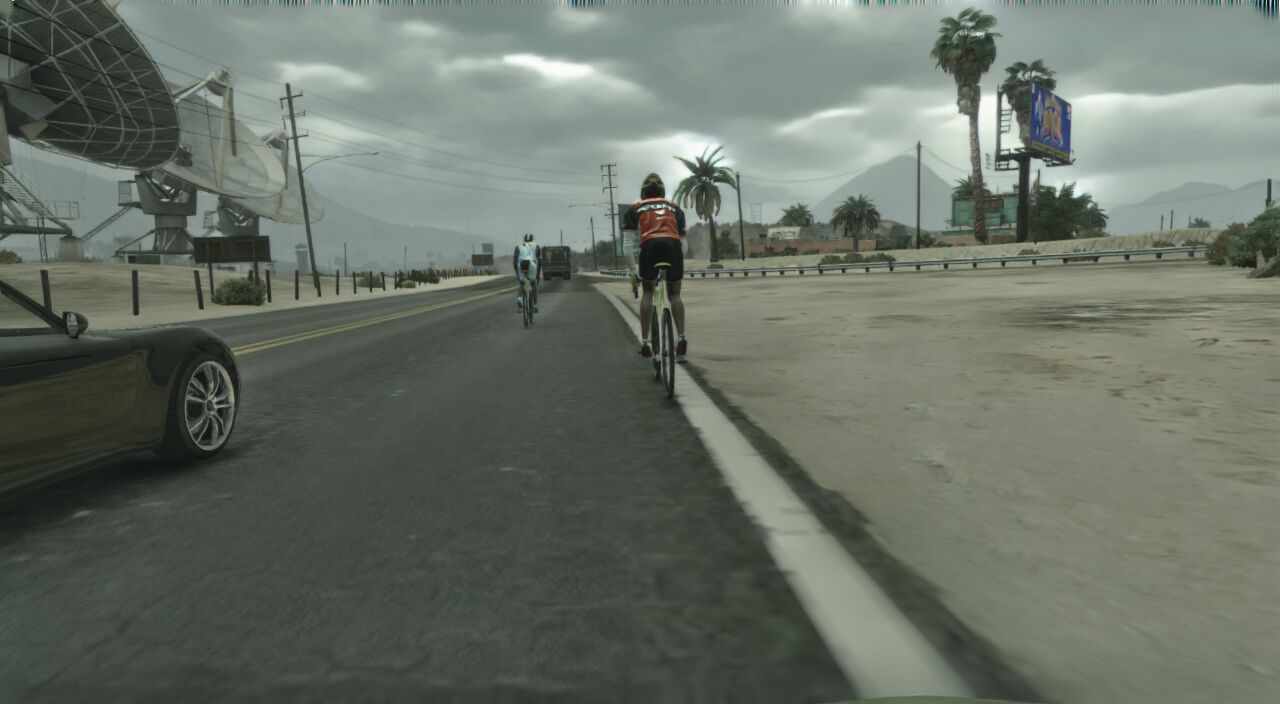}
\end{tabular}
\vspace{-3mm}
\caption{Qualitative comparison of different style-transfer methods and semantic consistency losses in the \textbf{GTA} $\rightarrow$ \textbf{Cityscapes} setting}
\label{fig:style-transfer-comparison}
\end{figure*}

\def\myim#1{\includegraphics[height=16.5mm]{img/quali/#1}}

\def\myimwboxi#1{\begin{tikzpicture}
    \node[] at (0,0) {\includegraphics[height=16.5mm]{img/quali/1/#1}};
    \draw [cyan, dashed, line width=0.5mm] (-1.6, -0.1) rectangle (-0.2, 0.8);
    \draw [cyan, dashed, line width=0.5mm] (1.35, 0.2) rectangle (1.6, 0.6);
\end{tikzpicture}}

\def\myimwboxii#1{\begin{tikzpicture}
    \node[] at (0,0) {\includegraphics[height=16.5mm]{img/quali/2/#1}};
    \draw [cyan, dashed, line width=0.5mm] (-1.1, -0.1) rectangle (-0.75, 0.8);
    \draw [cyan, dashed, line width=0.5mm] (0, -0.2) rectangle (1.5, 0.3);
\end{tikzpicture}}

\def\myimwboxiii#1{\begin{tikzpicture}
    \node[] at (0,0) {\includegraphics[height=16.5mm]{img/quali/3/#1}};
    \draw [cyan, dashed, line width=0.5mm] (0.3, -0.2) rectangle (1.5, 0.6);
\end{tikzpicture}}

\begin{figure*}
  \setlength\tabcolsep{.1 pt}
  \renewcommand{\arraystretch}{0.1} 
\centering
 \begin{tabular}{ccccc}

Target image
& Ground truth
    &ProCA \cite{proca}
    &SePiCo \cite{sepico}
    &\method(Ours)
\\ \myimwboxi{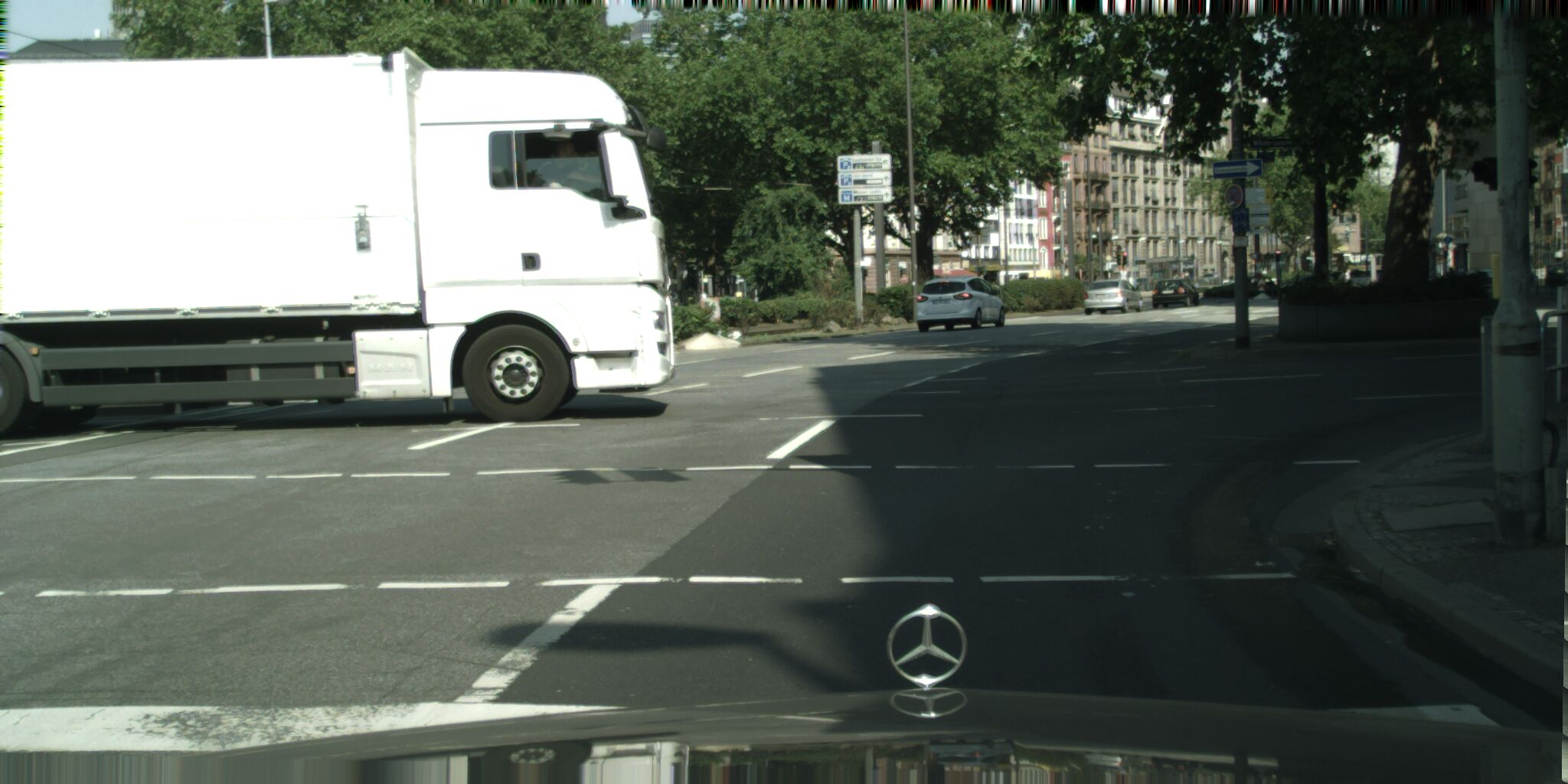}&
    \myimwboxi{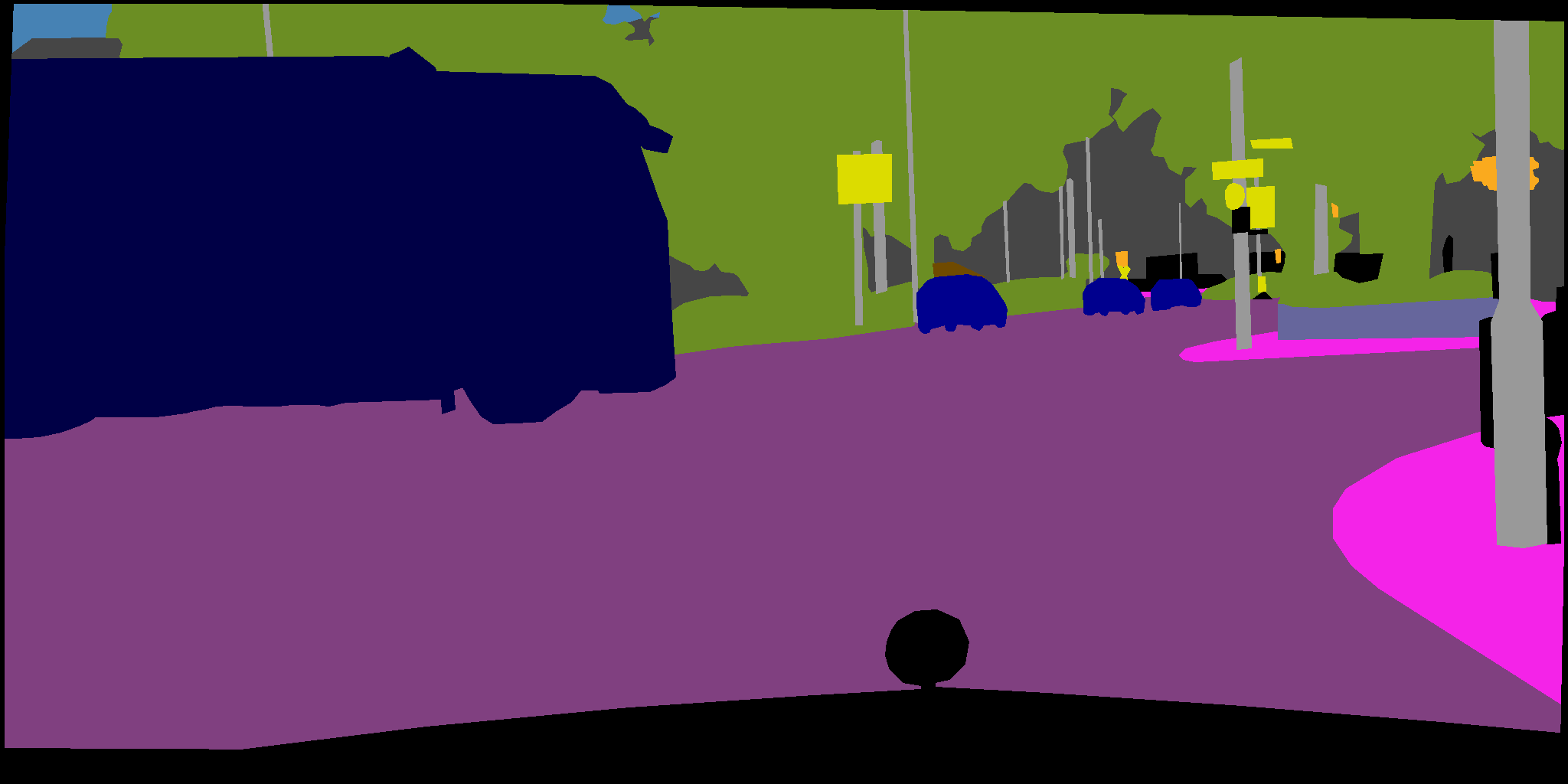}  &
    \myimwboxi{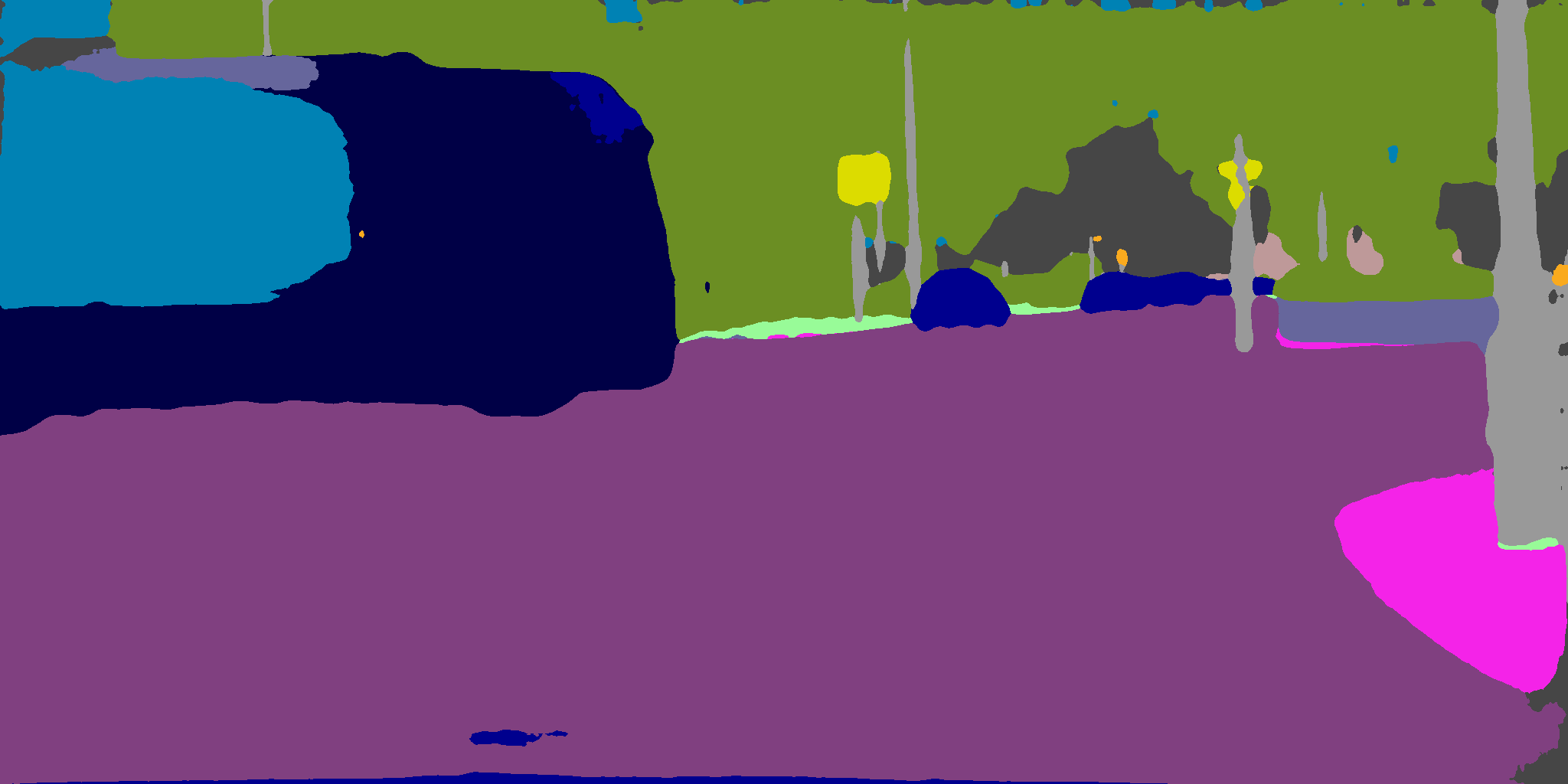}  &
    \myimwboxi{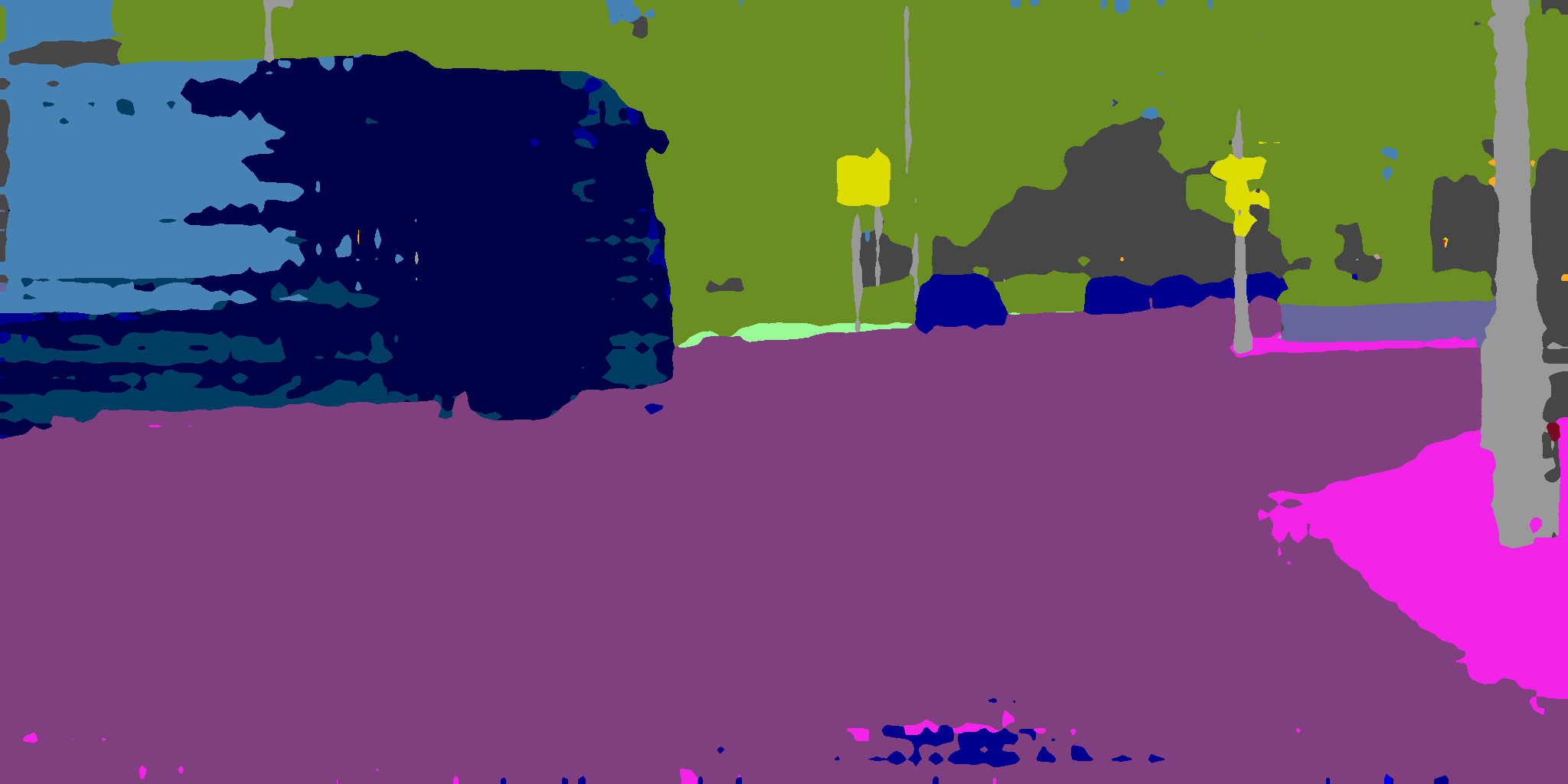} &
    \myimwboxi{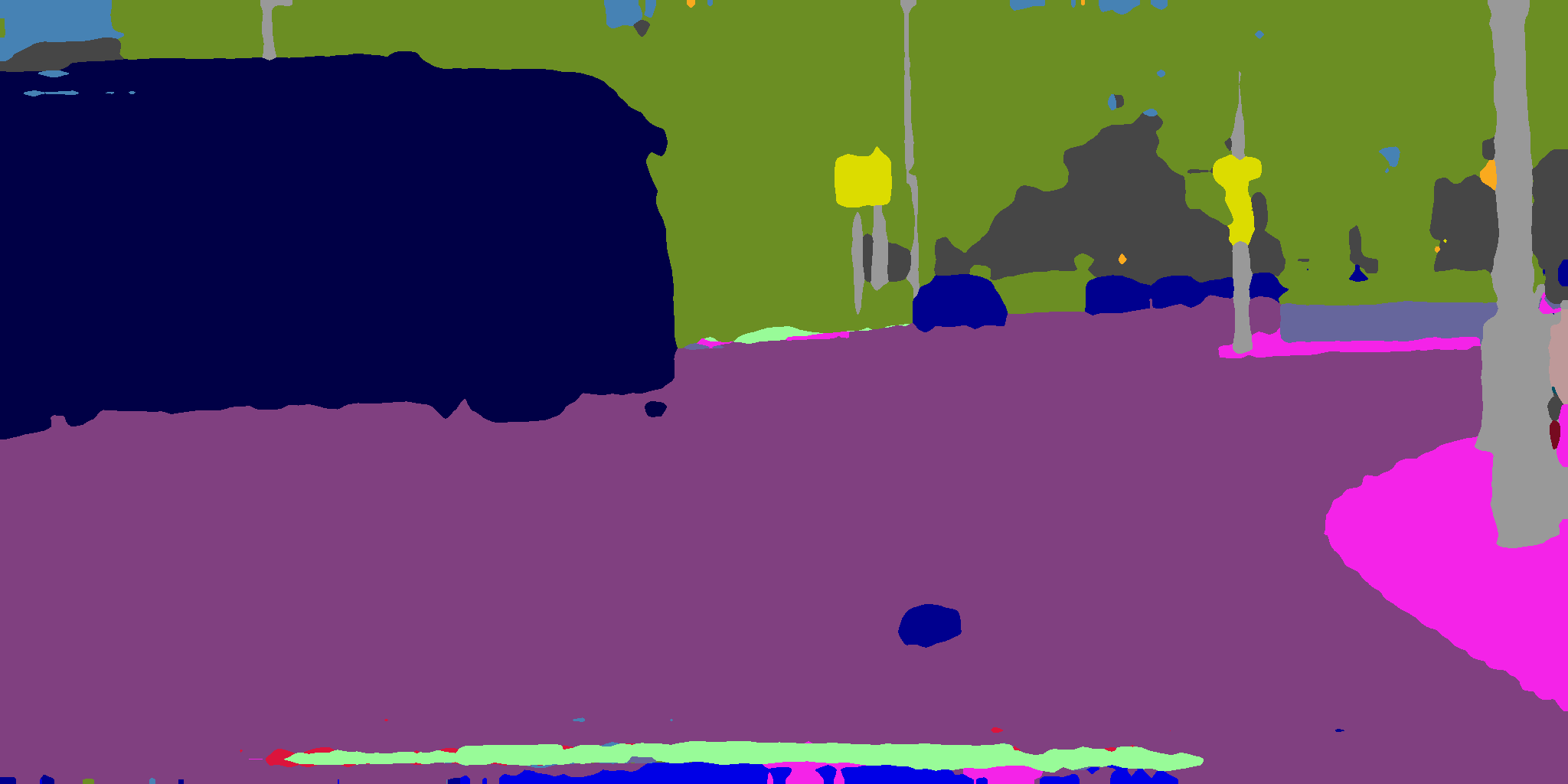}\\

    \myimwboxii{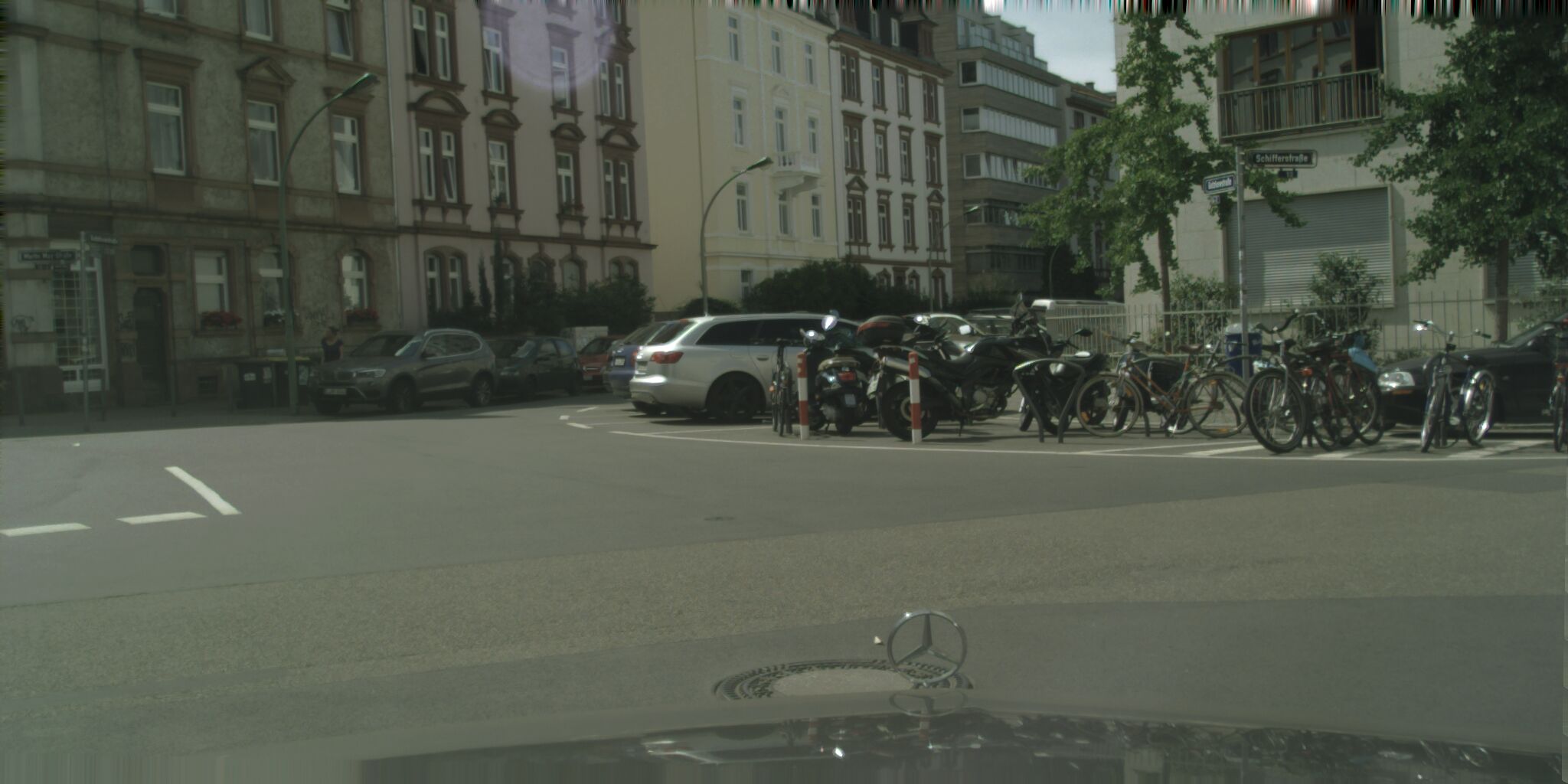} &
    \myimwboxii{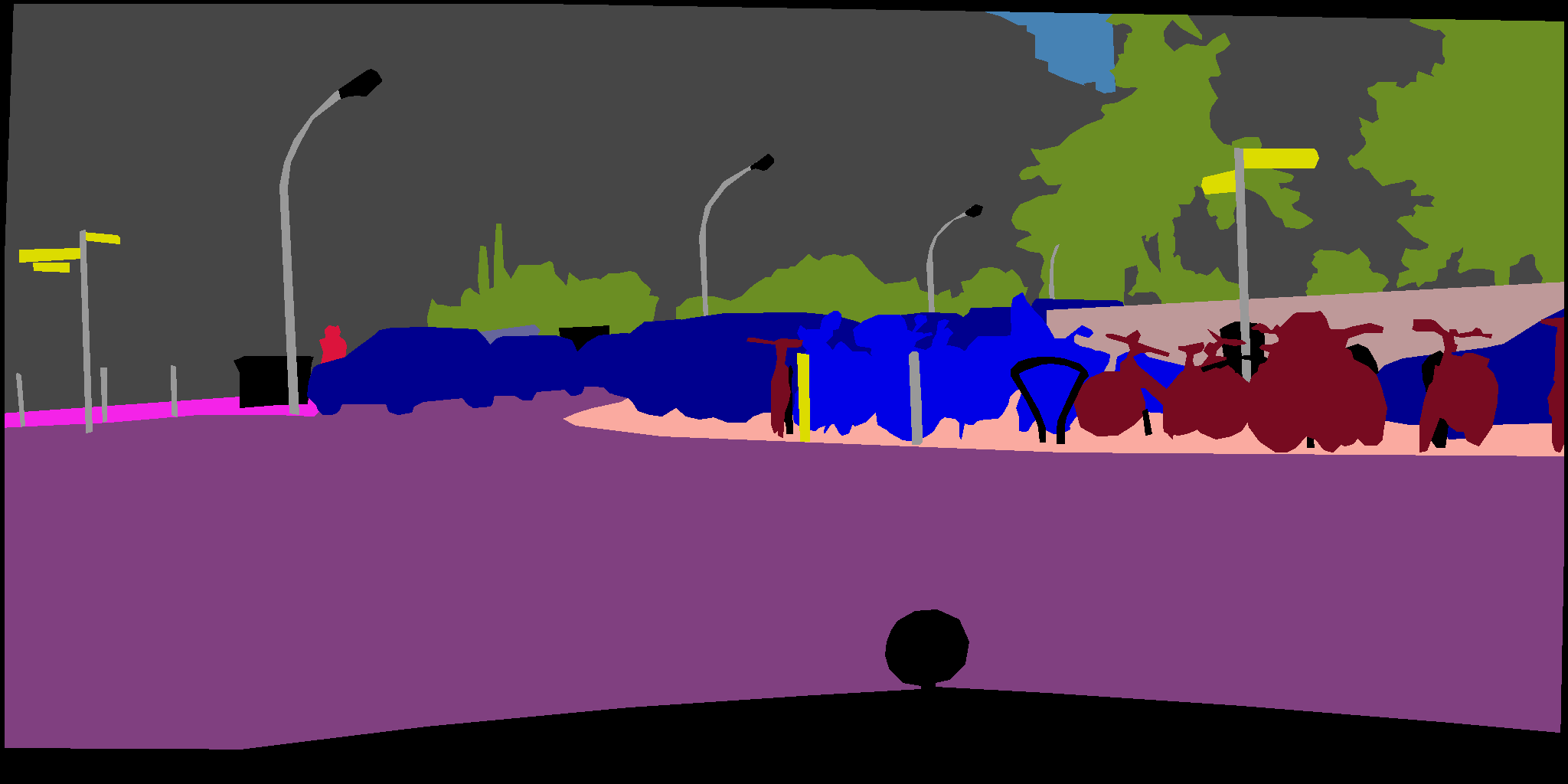}  &
    \myimwboxii{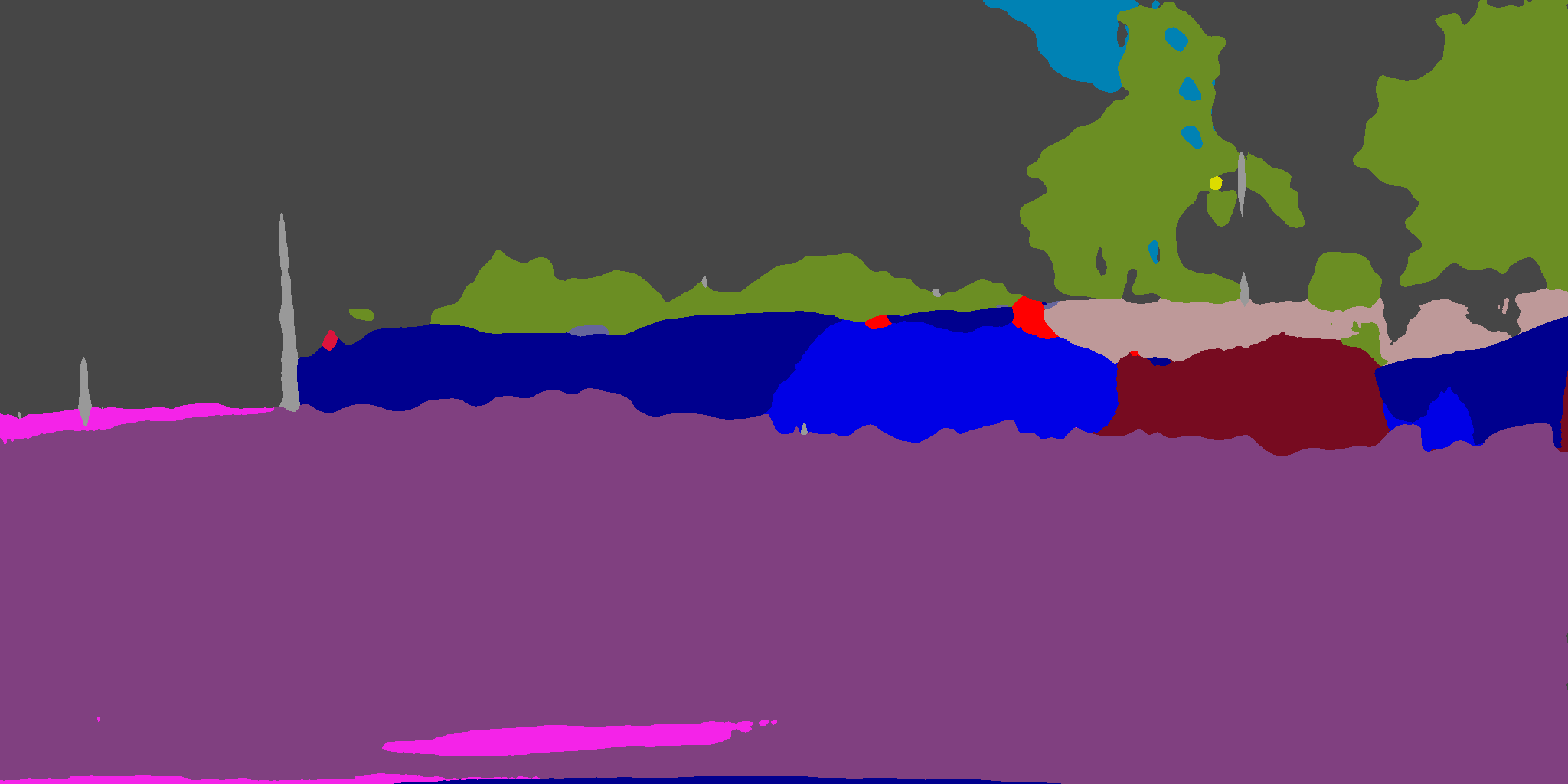}  &
    \myimwboxii{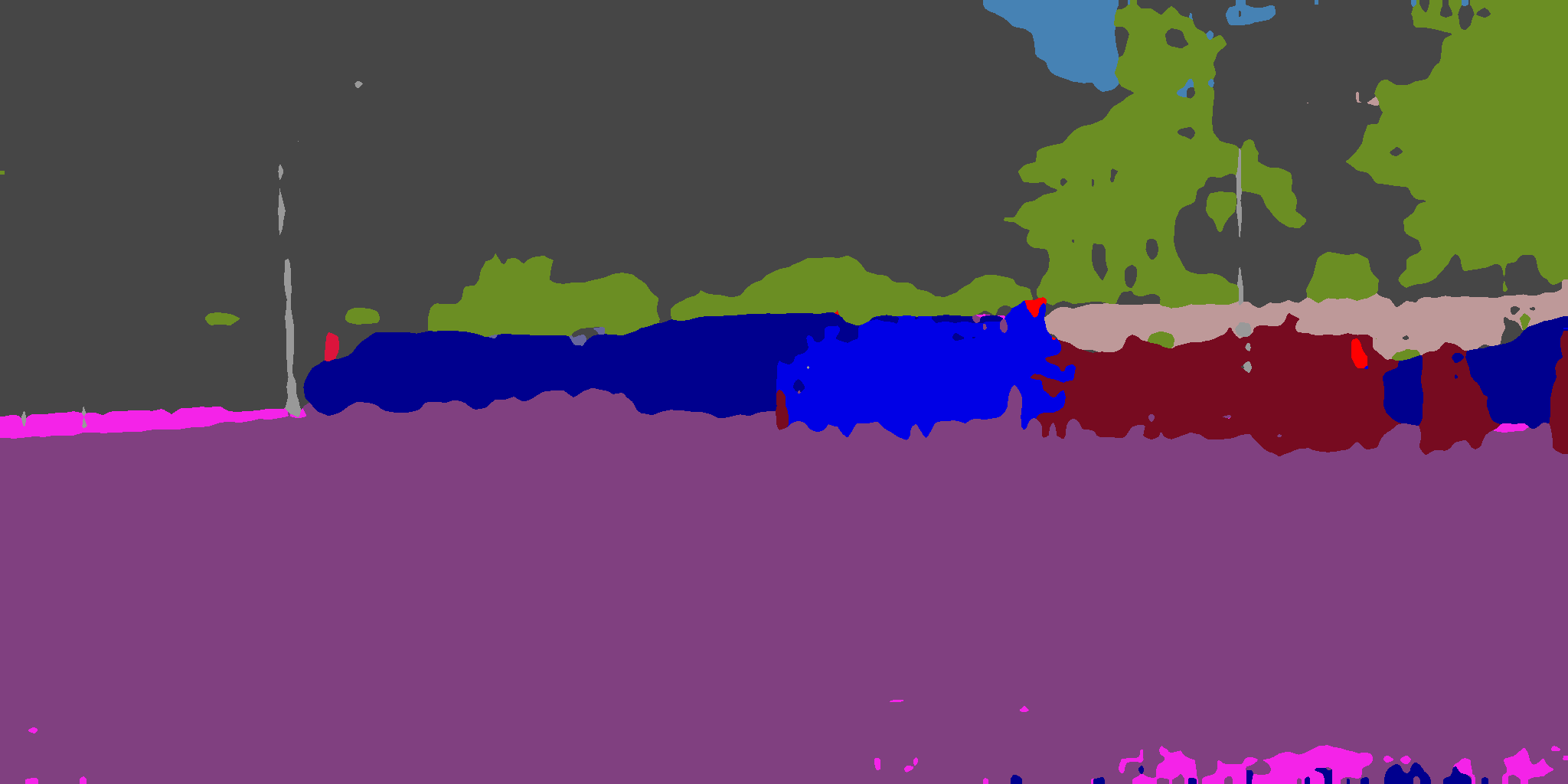} &
    \myimwboxii{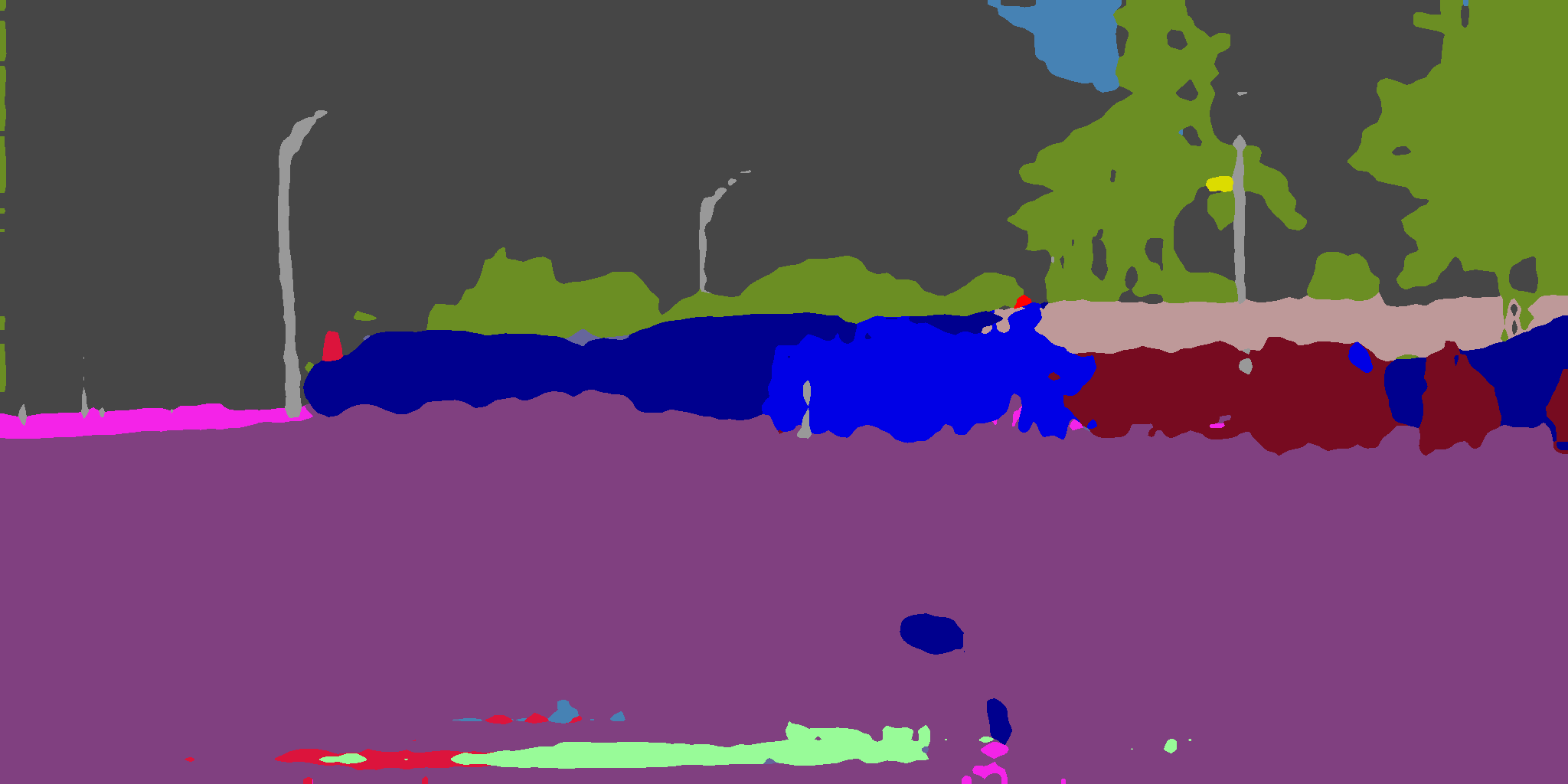}\\

    \myimwboxiii{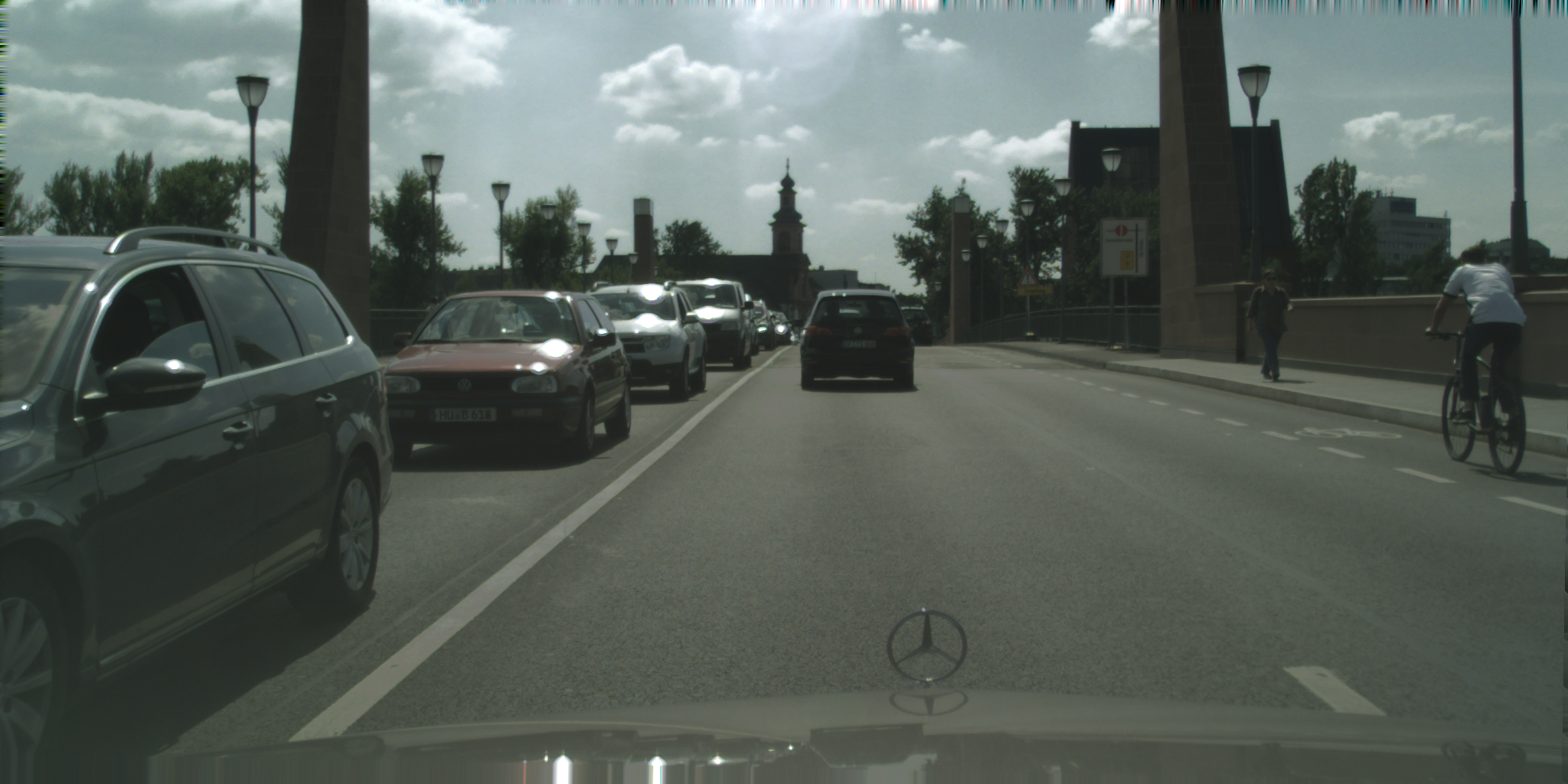} &
    \myimwboxiii{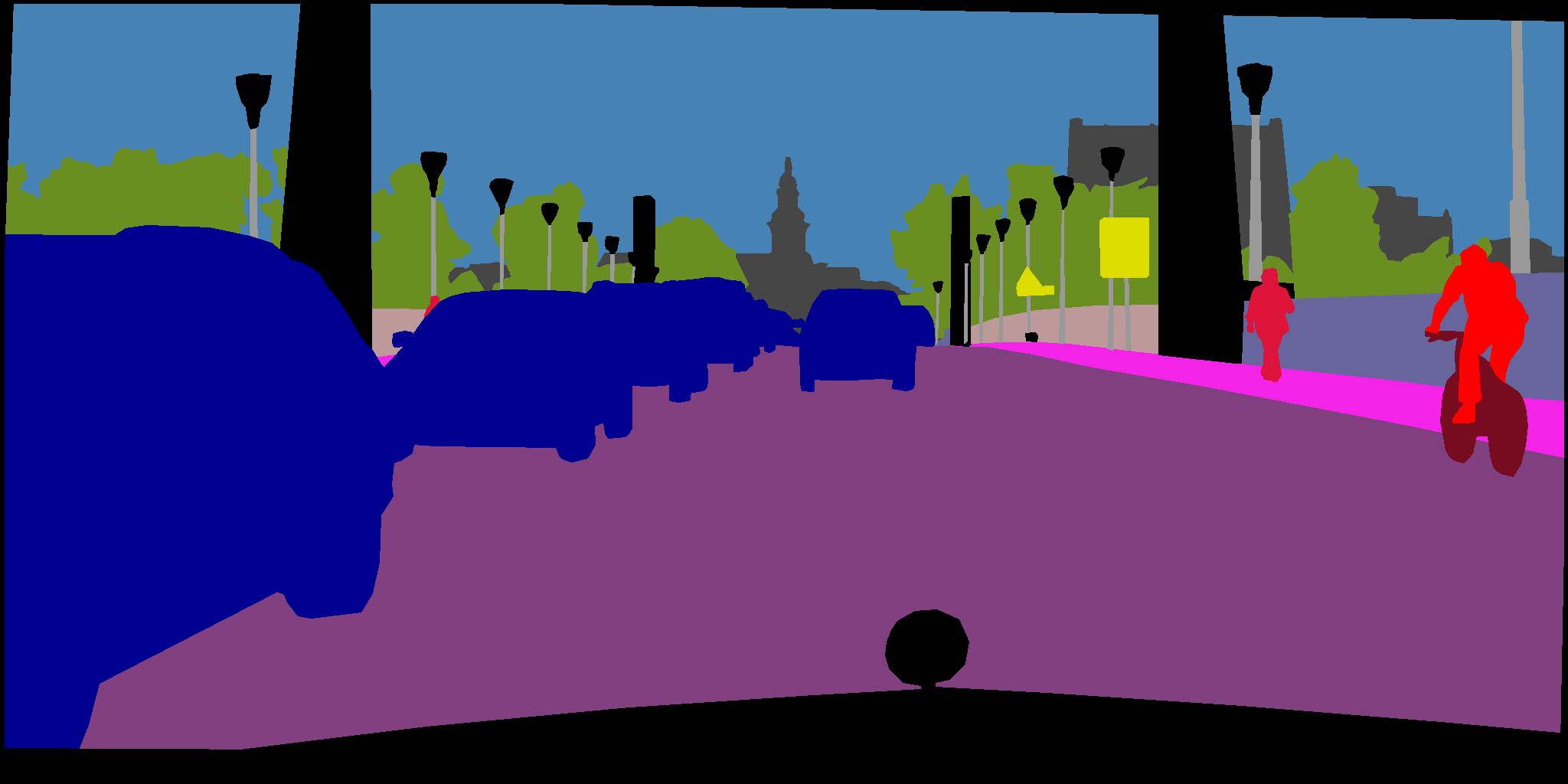}  &
    \myimwboxiii{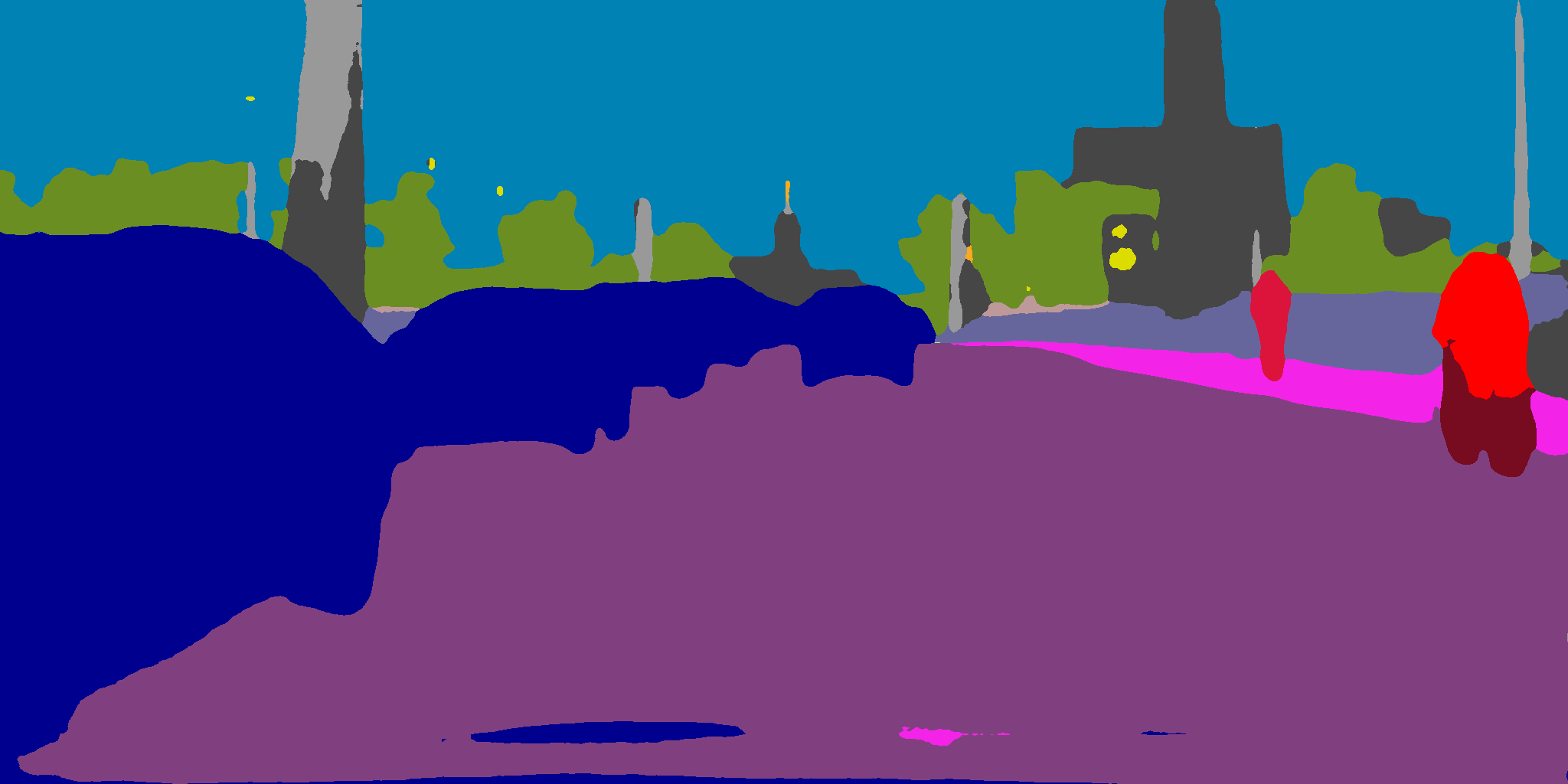}  &
    \myimwboxiii{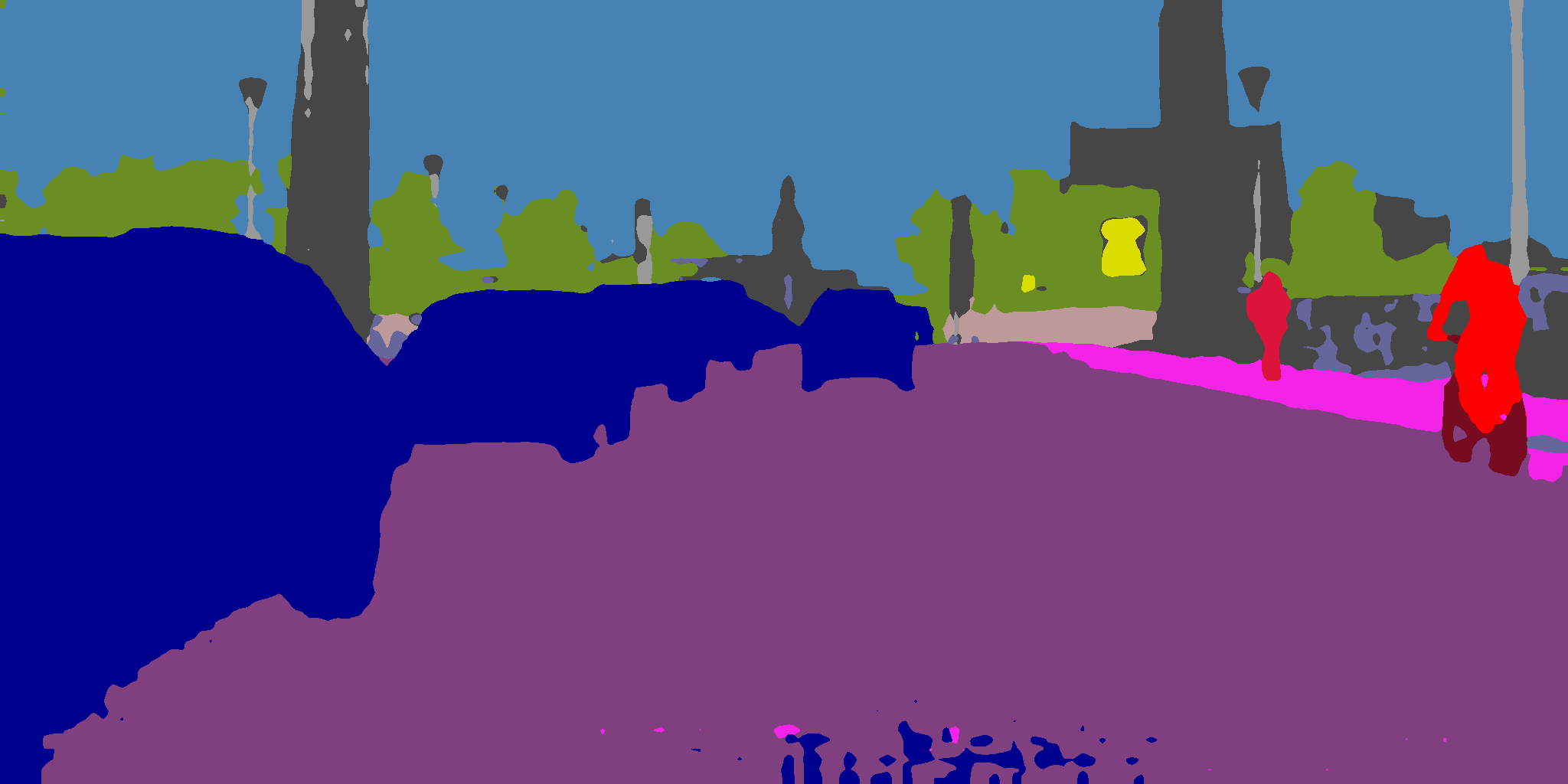} &
    \myimwboxiii{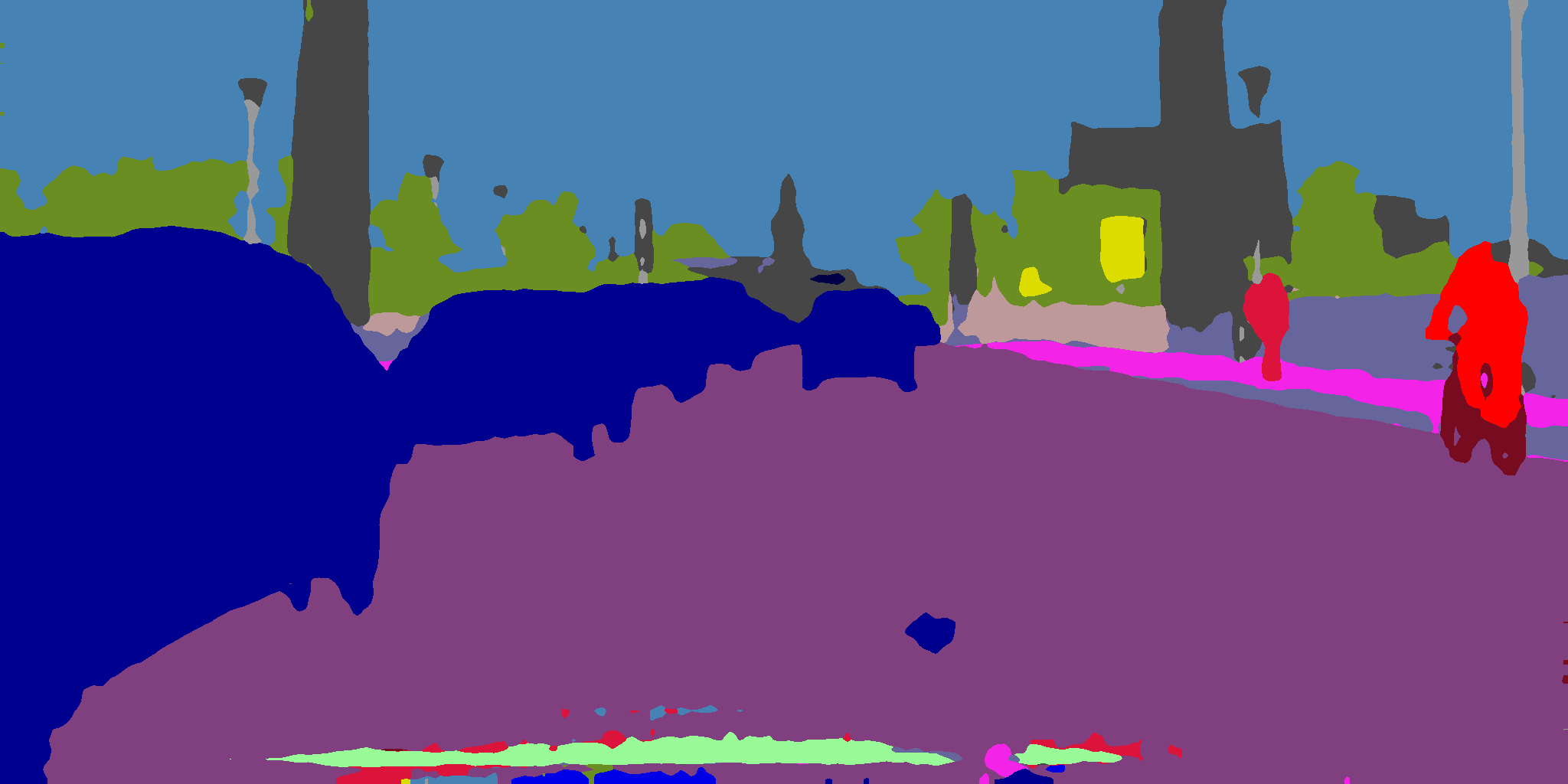}
\end{tabular}
\vspace{-3mm}
\caption{Qualitative comparison of the segmentation maps of \method with the state-of-the-art methods~\cite{proca,sepico} on \textbf{GTA} $\rightarrow$ \textbf{Cityscapes}}
\label{fig:quali}
\vspace{-3mm}
\end{figure*}

\noindent\textbf{Comparison with style transfer methods.}
\begin{table}[]
\centering
\small
\caption{Ablation on the \textbf{GTA} $\rightarrow$ \textbf{Cityscapes} benchmark. \textbf{Top}: Comparison with alternative style transfer methods. \textbf{Bottom}: Impact of the various semantic consistency losses}
\label{tab:abla_diffstl}
\resizebox{0.8\columnwidth}{!}{%
\begin{tabular}{ccccc@{}}
\toprule
 
&FDA~\cite{fda} & GPA~\cite{gpa} & AdaIN~\cite{adain} & Ours \\ 
\midrule
mIoU & 60.5 & 59.4 & 59.1 & \textbf{62.2}\\ 
\midrule
\midrule
& w/o $\mathcal{L}_\text{SC}$  & $\mathcal{L}_\text{CE}$  & MSE        & $\mathcal{L}_\text{SC}$ (Ours)  \\ 
 \midrule
 mIoU  & 58.0  & 58.6  & 61.6  & \textbf{62.2} \\ 
 \bottomrule
\end{tabular}}
\vspace{-5mm}
\end{table}
We now extend our comparison by evaluating alternative style-transfer methods to gauge the advantage of contrastive style transfer in \method. We consider the neural network-based AdaIN~\cite{adain} and training-free methods (\eg, FDA~\cite{fda}, GPA~\cite{gpa}). In these experiments, we replace the CUT module in our \method with each alternative method and employ the prototypical contrastive loss $\mathcal{L}_\text{PCL}$ in~\cref{eqn:pcl} as the training objective.
We report the quantitative results in the top half of~\cref{tab:abla_diffstl} and the qualitative results in~\cref{fig:style-transfer-comparison}. We can observe that our \method empirically outperforms all the aforementioned style transfer competitors. From this qualitative comparison, we observe that FDA or GPA generally generate image with poor quality and many artifacts. On the contrary, our method outputs images where the content of the source image is preserved but the style is well-transferred. The difference is  especially clear in the sky region where the most methods fail.

\noindent\textbf{Evaluation of the semantic consistency loss.}
We compare our proposed semantic consistency loss (\ie, $\mathcal{L}_\text{SC}$ in ~\cref{eqn:sem-cons-loss}) for pixel-level alignment with some other alternative losses from the \task literature. First, we consider a baseline without consistency loss (referred to as \textit{w/o} $\mathcal{L}_\text{SC}$). Then, we consider a baseline, referred to as $\mathcal{L}_\text{CE}$, which is inspired by CyCADA~\cite{cycada} and uses the cross-entropy between the estimated segmentations. Finally, we include a variant of \method where $\mathcal{L}_\text{SC}$ is replaced by a MSE loss.
The quantitative results in the bottom half of~\cref{tab:abla_diffstl} show that the last two variants that operate at the feature level clearly perform better. Among the two variants of \method, the contrastive formulation attains higher performance. The qualitative results are shown in the~\cref{fig:style-transfer-comparison}.

\begin{table}[]
\centering
\small
\setlength\tabcolsep{4. pt}
\renewcommand{\arraystretch}{0.4} 
\caption{Impact of design choices for prototypes estimation in terms of mIoU on the \textbf{GTA} $\rightarrow$ \textbf{Cityscapes} benchmark}
\label{tab:abla_estsrc}
\begin{tabular}{lcc|cc}
\toprule
Method & w/o CUT & CUT & w/o CAM & w/ CAM \\ 
\midrule
\multicolumn{1}{c}{w/o prototype} & 57.5 & 57.1 & - & -\\
\multicolumn{1}{c}{Source prototype} & 58.6  & 60.5 & - & -\\
\multicolumn{1}{c}{Mixed prototype}  & 60.0  & 62.2 & 60.1 & 62.2\\ 
\bottomrule
\end{tabular}%
\vspace{-3mm}
\end{table}

\noindent\textbf{Effect of mixed prototype estimation.}
We now analyse the design choices for prototype estimation.
The prototypes can be estimated from the original source images or from the mixed images obtained by ClassMix~\cite{dacs}. These two solutions are compared to a baseline where the prototypes are not used. We perform experiments with and without style transfer with CUT and report the results in the left of~\cref{tab:abla_estsrc}. First, we observe that estimating the prototypes with features from the mixed images leads to higher mIoUs. This result shows that including cross-domain information in the prototypes helps adaptation. Furthermore, when the mixed prototypes are used, combing with CUT further boosts the performance ($62.2\%$ vs $60.0\%$) showing that the prototypes with rich cross-domain information improve both pixel- and feature-level alignment.

\noindent\textbf{Effect of CAM-based weighting.}
We now validate our CAM-based solution to estimate the weighted prototypes (described in~\cref{sec:feat-align}). In the right of~\cref{tab:abla_estsrc}, we compare the performance of models where the prototypes are estimated with and without the proposed CAM-based weighting technique and trained with CUT. These results demonstrate the effectiveness of the weighted prototypes as they outperform the unweighted prototypes by $+1.1\%$ of mIoU.

\section{Conclusion}
\label{sec:con}
In this work, we presented \method, a novel approach that unifies feature-level and pixel-level cross-domain alignment. \method was integrated with the mean-teacher self-training framework and was shown to improve the performance of DASS through the joint use of prototypical contrastive loss and style transfer. On one hand, our jointly trained style transfer module produced high-quality stylized images that bridge the domain gap and aid in prototype estimation. On the other hand, the well-estimated prototypes, in conjunction with the prototypical contrastive loss, further reinforce the feature-level alignment and enhance the DASS performance. Both quantitative and qualitative experiments demonstrated the effectiveness of \method, outperforming existing state-of-the-art methods.

\noindent\textbf{Acknowledgements.}
{\small This paper has been supported by the French National Research
Agency (ANR) in the framework of its JCJC program (Odace, project ANR-20-CE23-0027) and NSFC (62176155), Shanghai Municipal Science and Technology Major Project (2021SHZDZX0102).

}
\vfill\pagebreak

{\small
\bibliographystyle{ieee_fullname}
\bibliography{egbib}
}

\clearpage
\twocolumn[\vspace*{2em}\centering\Large\bf%
{\Large Supplementary Material}%
\vspace*{2em}]

\setcounter{table}{0}
\renewcommand{\thetable}{A\arabic{table}}%
\setcounter{figure}{0}
\renewcommand{\thefigure}{A\arabic{figure}}%
\setcounter{equation}{0}
\renewcommand{\theequation}{A\arabic{equation}}%

\appendix

The supplementary material is organized as follows: in Sec.~\ref{sec:app-additional-losses} we detail all the losses used for training our method. In Sec.~\ref{sec:app-qualitative} we provide additional discussion on the qualitative results.

\section{Additional Training Details}
\label{sec:app-additional-losses}

Due to lack of space we provide additional training losses that are used to train \method in the supplementary material. The weighted prototype estimation (described in Sec.~\ref{sec:feat-align} of the main paper) is summarized in Algo.~\ref{alg:wpe}.

\paragraph{Class Activation Maps.} To recall, in order to obtain the class activation maps (CAM) we learn the linear layer $\mathbf{w} \in \mathbb{R}^{K \times D}$, where $K$ and $D$ are the number of classes and channel dimension of the latent features, respectively. The predicted probability scores are obtained by first applying global average pooling on the features obtained from the backbone $f_b$, followed by projecting them with the linear layer as:

\begin{equation}
    \hat{p}_c = \sigma(\frac{1}{H'W'}\displaystyle \sum^{D}_{d=1} \mathbf{w}_{c, d} \displaystyle \sum^{H' \times W'}_{j=1} \mathbf{f}_{d, j})
\end{equation}
where $\sigma(\cdot)$ is the sigmoid function.

Given, objects from multiple classes can be present in an image, we use a multi-label classification loss to train the network. In details, we construct the binary labels $y \in \{0, 1\}^K$, from the ground truth labels of the source and the pseudo-labels of the target domain, which indicate the existence of a class in an image. The network is then trained with a standard binary cross-entropy (BCE) loss as:

\begin{equation}
\label{eqn:bce-loss-cam}
    \mathcal{L}_\text{CAM} = - \frac{1}{K} \displaystyle \sum^{K}_{c=1} y_c \log \hat{p}_c + (1 - y_c) \log (1 - \hat{p}_c)
\end{equation}

\paragraph{Diversity Regularization Loss.} The prototypical contrastive loss $\mathcal{L}_\text{PCL}$ (described in Sec.~\ref{sec:feat-align}) can be prone to trivial solution where all the features are assigned to a single prototype. This can happen because of the absence of labels in the target domain and the possibly noisy and erroneous pseudo-label, especially in the presence of a learnable projection. We thus follow \cite{sepico}, and apply a regularization term to guarantee feature diversity:
\begin{equation}
\label{eqn:diversity-loss}
    \mathcal{L}_\text{Reg} = \frac{1}{K\log K}\sum_{c=1}^K \frac{\exp(\overline{\mathbf{v}} \cdot \mathbf{p}_c / T)}{\sum_{k=1}^K \exp(\overline{\mathbf{v}}\cdot \mathbf{p}_k/T)}
\end{equation}
where $\overline{\mathbf{v}} = \frac{1}{|B||H'W'|}\sum^{B}_{i=1} \sum^{H' \times W'}_{j=1} \mathbf{v}_{i,j}$ is the mean feature computed from a mini-batch of size $|B|$ and $\mathbf{v}_{i,j}$ is the projected features from the $j^\text{th}$ pixel location in the $i^\text{th}$ image.

\renewcommand{\algorithmicrequire}{\textbf{Input:}}
\renewcommand{\algorithmicensure}{\textbf{Output:}}
\begin{algorithm}[!h]
\caption{Weighted prototype estimation}\label{alg:wpe}
\begin{algorithmic}
\Require CAM $\mathbf{M}$, projected features $\mathbf{v}_j$, set of pixel locations $\mathcal{N}$, semantic categories $K$, momentum update parameter $\gamma$
\Ensure Prototypes $\{\mathbf{p}_c\}_{c=1}^K$
\For {c=1:$K$}
    \State Select features $\mathbf{v}_j$ belongs to class $c$
    \State Select $N$ features with highest CAM score
    \State $\mathbf{p}_c' = \frac{\sum_{j \in \mathcal{N}_c} \mathbf{M}_{c,j} \mathbf{v}_j}{\sum_{j' \in \mathcal{N}_c} \mathbf{M}_{c,j'}}$
    \State $\mathbf{p}_c \leftarrow \gamma \mathbf{p}_c + (1-\gamma)\mathbf{p}_c'$
    \State $\mathbf{p}_c \leftarrow \frac{\mathbf{p}_c}{\|\mathbf{p}_c\|_2}$
\EndFor
\State \Return $\{\mathbf{p}_c\}_{c=1}^K$
\end{algorithmic}
\end{algorithm}

\paragraph{Style Transfer Losses.} Besides the patch-based NCE loss for style transfer (described in Sec.~\ref{sec:pixel-align}), we also use the adversarial loss to train the generator $\mathcal{G}$ and the discriminator $\mathcal{E}$. The adversarial loss encourages the stylized source images to be visually similar to the target distribution, and is given as:

\begin{equation}
\label{eqn:gan-loss}
\begin{split}
    \mathcal{L}_\text{GAN} &= \mathbb{E}_{X^\mathrm{T} \sim \mathcal{D}_\mathrm{T}} \log \mathcal{E}(X^\mathrm{T}) \\
    &+ \mathbb{E}_{X^\mathrm{S} \sim \mathcal{D}_\mathrm{S}} (1 - \log \mathcal{E}(\mathcal{G}(X^\mathrm{S})))
\end{split}
\end{equation}

\paragraph{CUT Architecture.} The CUT \cite{cut} adopts an encoder-decoder architecture for its generator $\mathcal{G}$. The first half of $\mathcal{G}$ is regarded as the encoder $\mathcal{G}_{enc}$ and the second half is the decoder $\mathcal{G}_{dec}$. Thus, the generator is the composition $\mathcal{G}_{enc} \circ \mathcal{G}_{dec}$.

Given an image $X^\mathrm{S}$, the $L$ layers features are extracted from $\mathcal{G}_{enc}$ for computing $\mathcal{L}_{PatchNCE}$. Naturally, the features from the deeper layers in the encoder corresponds to a larger patch due to the increasing receptive field. At layer $l$, the features $\{\mathbf{z}_{i,l}\}_{i=1}^N$ is produced as:
\begin{equation}
    \mathbf{z}_{i,l}^{S} = H_l(\mathcal{G}_{enc,l}(X^\mathrm{S}))_i
\end{equation}

where $H_l$ is a 2-layer MLP followed by a $\ell_2$ normalization. Similarly, the stylized features $\{\hat{\mathbf{z}}_{i,l}^{S\rightarrow ^\mathrm{T}}\}_{i=1}^N$ is obtained in the same manner with the stylized image $\hat{X}^{S\rightarrow ^\mathrm{T}} = \mathcal{G}(X^\mathrm{S})$.

The consistency in the translation is guarantee by the $\mathcal{L}_{PatchNCE}$ computed with properly chosen positive and negative pairs. Since the patches in the same location should have similarity before and after the stylization, for $\hat{\mathbf{z}}_{i,l}^{S\rightarrow ^\mathrm{T}}$, the positive patch $\mathbf{z}^\mathrm{S}_{i,l,+}\ = \mathbf{z}^\mathrm{S}_i$and the negative patches $\{\mathbf{z}^\mathrm{S}_{i,l,-}\} = \{\mathbf{z}_{k,l}^\mathrm{S}\}_{k\in \{1\dots N\} / i}$. Finally, the PatchNCE loss is computed as:
\begin{equation}
\mathcal{L}_{PatchNCE} = \sum_{l=1}^L \sum_{i=1}^N \ell_{PatchNCE}(\hat{\mathbf{z}}_{i,l}^{\mathrm{S}\rightarrow \mathrm{T}}, \mathbf{z}^\mathrm{S}_{i,l,+}, \{\mathbf{z}^\mathrm{S}_{i,l,-}\})
\end{equation}

\paragraph{Overall Training Objective.} The final training objective is given as:

\begin{equation}
    \label{eqn:total-losses}
    \begin{split}
        \mathcal{L}_\text{Joint}  =& \mathcal{L}_\text{CE} + \lambda_\text{PCL} \mathcal{L}_\text{PCL} + \lambda_\text{CAM} \mathcal{L}_\text{CAM} +
         \lambda_\text{Reg} \mathcal{L}_\text{Reg} + \\
         & \lambda_\text{style} (\mathcal{L}_\text{GAN} + \mathcal{L}_\text{PatchNCE} + \mathcal{L}_\text{SC})
    \end{split}
\end{equation}
where $\lambda_\text{PCL}$, $\lambda_\text{CAM}$, $\lambda_\text{Reg}$, $\lambda_\text{style}$ are the weighing factors.

\section{Qualitative Analysis}
\label{sec:app-qualitative}

As shown in Fig.~\ref{fig:quali} our method performs better on several difficult situations than \cite{proca} and \cite{sepico}. Our method better classify the same objects in the same category correctly, such as the truck in the first row and the fence and wall in the third row. \method also outperforms other methods on small and fine objects, for instance the poles and traffic lights in the second row. In particular, \method is able to distinguish better similar objects, for instance, in the example of the first row, our methods classified correctly the truct while other methods confused it with the sky. Another example is shown in the second row, the bikes and motorbikes are easy to be confused, yet \method gives a better segmentation at the junction of two similar objects.

\end{document}